\documentclass{article}

\PassOptionsToPackage{numbers, compress}{natbib}
    \usepackage[preprint]{main}

\usepackage[utf8]{inputenc} % allow utf-8 input
\usepackage[T1]{fontenc}    % use 8-bit T1 fonts
\usepackage[colorlinks=true,breaklinks=true,bookmarks=false,citecolor=green]{hyperref}       % hyperlinks
\usepackage{url}            % simple URL typesetting
\usepackage{booktabs}       % professional-quality tables
\usepackage{amsfonts}       % blackboard math symbols
\usepackage{nicefrac}       % compact symbols for 1/2, etc.
\usepackage{microtype}      % microtypography
\usepackage{xcolor}         % colors
\usepackage{graphicx}
\usepackage{amsmath}
\usepackage{amssymb}
\usepackage{subcaption}
\usepackage{wrapfig}
\usepackage{pifont}
\usepackage{algorithmic}
\usepackage[ruled]{algorithm2e}
\usepackage{multirow}
\usepackage{fontawesome}% colors
\usepackage{cleveref}    
\usepackage{tikz}
\usepackage{xspace}

\newcommand{\ours}{Puppeteer}
\newcommand{\ourdata}{Articulation-XL2.0}
\newcommand{\res}{ModelsResource}
\newcommand{\boldstartspace}[1]{\medskip\noindent\textbf{#1}}

\newcommand{\purplecircle}[2][11pt]{%
  \tikz[baseline=(X.base)]\node[
    circle,
    draw=violet!20,
    fill=violet!20, 
    minimum size=#1, 
    inner sep=0pt,
    text=black,       
    font=\footnotesize  
  ] (X) {#2};%
}

\title{
  \raisebox{-0.4em}{\includegraphics[height=1.5em]{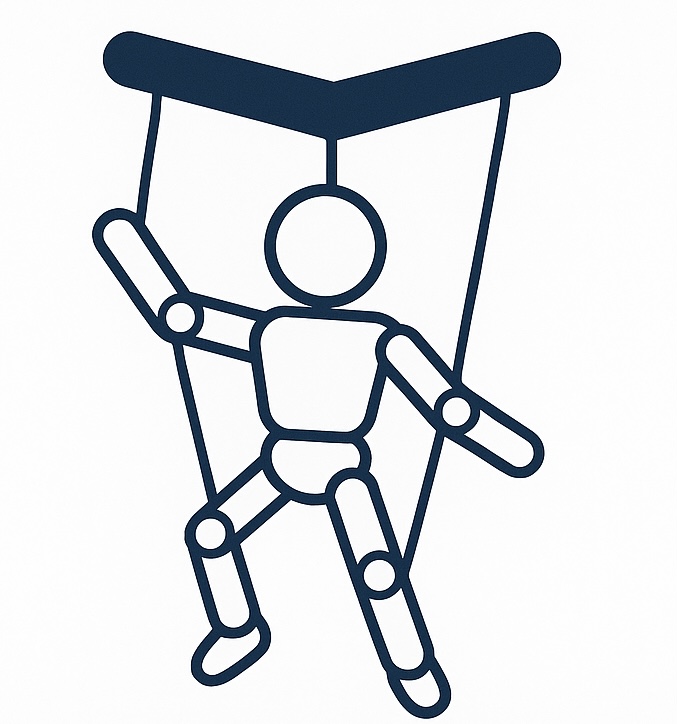}\hspace{0.1em}}
  Puppeteer: Rig and Animate Your 3D Models
}

\author{
    Chaoyue Song\textsuperscript{1,2}, 
    Xiu Li\textsuperscript{2}, 
    Fan Yang\textsuperscript{1}, 
    Zhongcong Xu\textsuperscript{2}, 
    Jiacheng Wei\textsuperscript{1}, \\
    \textbf{Fayao Liu\textsuperscript{3}, 
    Jiashi Feng\textsuperscript{2}, 
    Guosheng Lin$^{\dag}$\textsuperscript{1},
    Jianfeng Zhang$^{\dag}$\textsuperscript{2}} \\
    \textsuperscript{1}{Nanyang Technological University} \quad
    \textsuperscript{2}{ByteDance Seed} \\ 
    \textsuperscript{3}{Institute for Infocomm Research, A*STAR}\\
    {\tt\small \url{https://chaoyuesong.github.io/Puppeteer}}
    \vspace{-.20in}
}

\begin{document}

\makeatletter
\let\@oldmaketitle\@maketitle
\renewcommand{\@maketitle}{\@oldmaketitle
 \centering
    \includegraphics[width=0.95\linewidth]{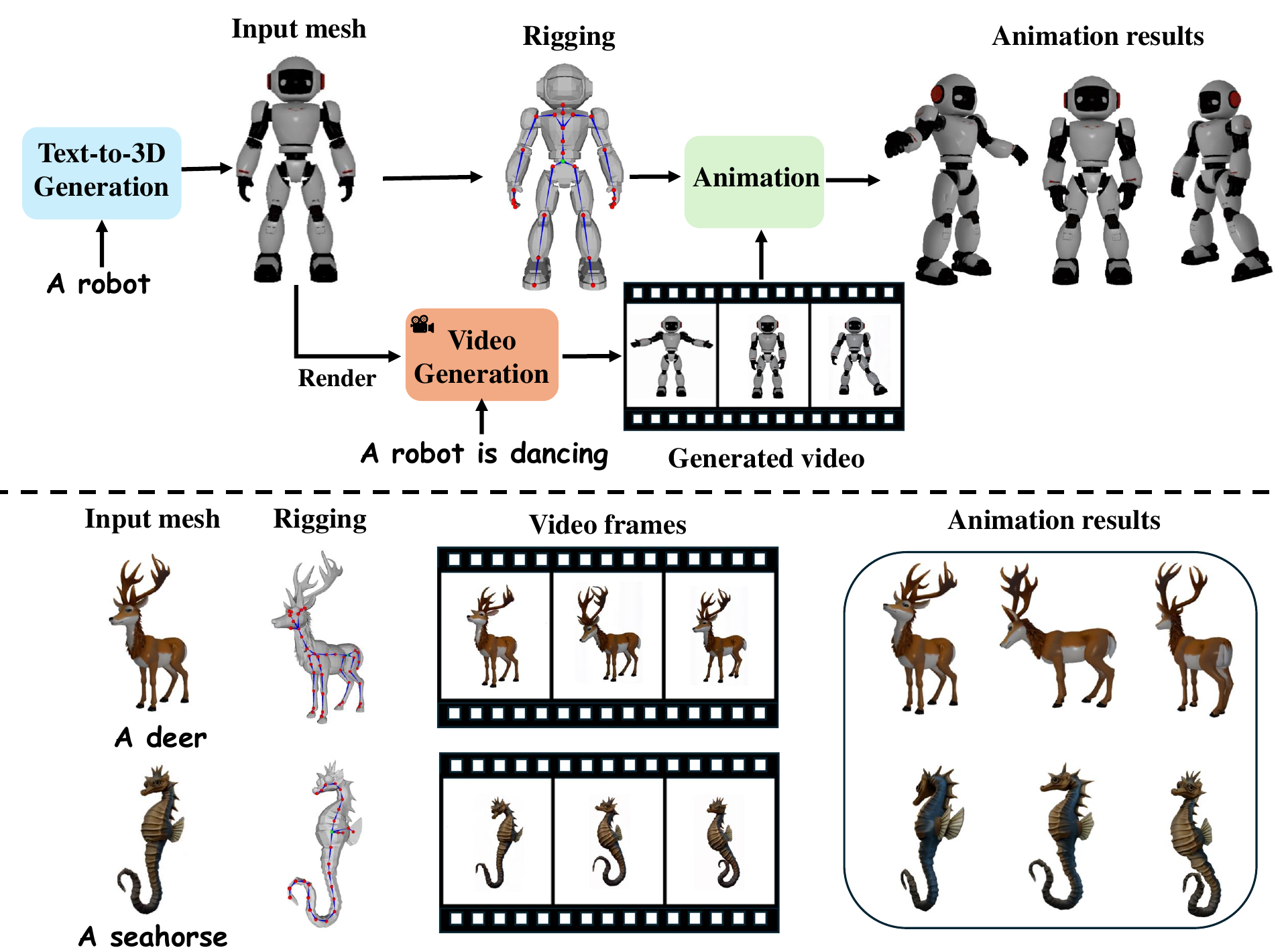}
  \captionof{figure}{
  Given a 3D model, we apply automatic rigging to create a skeleton structure with skinning weights. The input mesh is then rendered as input for video generation models \cite{klingai, jimengai2025}. Finally, we produce animations guided by the generated videos. The input 3D models are generated by \cite{hunyuan3d22025tencent}.}
  \label{teaser}
  \bigskip}
\makeatother

\maketitle

% 9 pages
\let\oldthefootnote\thefootnote
\let\thefootnote\relax
\footnotetext{$^{\dag}$ Corresponding authors. Email: chaoyue002@e.ntu.edu.sg.}
\let\thefootnote\oldthefootnote

\begin{abstract}
Modern interactive applications increasingly demand dynamic 3D content, yet the transformation of static 3D models into animated assets constitutes a significant bottleneck in content creation pipelines.  While recent advances in generative AI have revolutionized static 3D model creation, rigging and animation continue to depend heavily on expert intervention. We present \textbf{Puppeteer}, a comprehensive framework that addresses both automatic rigging and animation for diverse 3D objects. 
Our system first predicts plausible skeletal structures via an auto-regressive transformer that introduces a joint-based tokenization strategy for compact representation and a hierarchical ordering methodology with stochastic perturbation that enhances bidirectional learning capabilities. It then infers skinning weights via an attention-based architecture incorporating topology-aware joint attention that explicitly encodes inter-joint relationships based on skeletal graph distances. 
Finally, we complement these rigging advances with a differentiable optimization-based animation pipeline that generates stable, high-fidelity animations while being computationally more efficient than existing approaches.
Extensive evaluations across multiple benchmarks demonstrate that our method significantly outperforms state-of-the-art techniques in both skeletal prediction accuracy and skinning quality. The system robustly processes diverse 3D content, ranging from professionally designed game assets to AI-generated shapes, producing temporally coherent animations that eliminate the jittering issues common in existing methods.
\end{abstract}

\section{Introduction}
\label{sec:intro}

From AAA games and animated films to VR/AR experiences and robotic simulations, modern interactive media demands dynamic 3D content. While recent generative AI advances have accelerated the creation of high-fidelity 3D models with intricate geometry and textures, these assets remain predominantly static. Transforming static 3D models into animated versions requires two expert-driven processes: rigging (skeleton setup and skinning weight assignment) and animation. This manual, time-intensive workflow now constitutes a significant impediment to the efficiency of modern content creation pipelines.

Research communities have invested considerable effort in automating the rigging process. Early template-based techniques such as Pinocchio \cite{baran2007automatic} fit predefined skeletal structures to input meshes, achieving satisfactory results on specific categories but failing to generalize to arbitrary shapes. Template-free algorithms \cite{huang2013l1, au2008skeleton, cao2010point, lin2021point2skeleton,tagliasacchi2012mean} extract skeletal structures directly from geometric properties but frequently produce excessively dense or topologically incompatible joint configurations unsuitable for practical animation workflows. Deep learning approaches have substantially advanced the field: RigNet \cite{xu2020rignet} pioneered direct skeleton and skinning weight prediction from input shapes using graph neural networks, while MagicArticulate \cite{song2025magicarticulate} reformulated skeleton generation as an auto-regressive problem and introduced a large-scale dataset with detailed rigging annotations. Despite these innovations, significant challenges persist: RigNet struggles with complex mesh topologies due to its reliance on carefully crafted features and restrictive orientation requirements. MagicArticulate suffers from computational inefficiency during inference and limited generalization in its functional diffusion process for skinning weight prediction. 
Critically, both approaches address only the rigging stage of the pipeline, leaving the equally challenging animation process as a separate manual task that requires substantial expertise. 

In this work, we present \textbf{\ours{}}, a comprehensive framework that integrates automatic rigging and animation into a unified pipeline. 
To address the data scarcity and limited pose diversity in existing datasets, we expand the Articulation-XL dataset \cite{song2025magicarticulate} to 59.4k rigged models, including a carefully curated subset of 11.4k diverse pose examples that enhance generalization to varied pose inputs. This expanded dataset serves as the foundation for our learning-based approach.
To overcome the limitations of existing rigging approaches in handling diverse shapes and complex topologies, our system introduces key improvements to both fundamental rigging components. For skeleton generation, we employ auto-regressive transformers featuring joint-based tokenization and hierarchical sequence ordering with randomization, creating more compact representations while generating structurally coherent skeletons free from template dependencies. For skinning weight prediction, we propose an attention-based architecture incorporating topology-aware joint attention that explicitly encodes skeletal graph structure, achieving robust weight prediction with enhanced generalization and computational efficiency. Beyond rigging, we address the automatic animation challenge that previous methods have largely overlooked.
We introduce a differentiable optimization-based method that requires no neural network parameters yet produces stable, high-quality animations by combining our generated rigging with reference video guidance easily obtained from off-the-shelf video generation models. Our unified framework enables full automation from static meshes to animated assets, transforming the labor-intensive manual workflow into an efficient, accessible pipeline for diverse 3D content creation.

Extensive evaluations demonstrate the effectiveness of our approach across both rigging and animation tasks. For rigging, experiments on the expanded \ourdata{} dataset and ModelsResource benchmark \cite{ModelsResource2019, xu2019predicting} show significant improvements over state-of-the-art methods in skeleton accuracy and skinning weight quality. The robustness of our approach is further validated through successful application to diverse 3D content—from professionally designed game assets to AI-synthesized geometries. For animation, direct comparisons against recent 4D generation techniques \cite{uzolas2025motiondreamer, ren2024l4gm} show that our optimization-based approach produces more temporally consistent and visually faithful results while maintaining computational efficiency. Notably, our method eliminates the jittering artifacts commonly seen in learning-based approaches during complex motion sequences. The clean and stable animation results also highlight the reliability of our automatically generated rigs.

In summary, our work advances automated 3D model rigging and animation through four key contributions:
(1) An expanded large-scale articulation dataset with 59.4k rigged models including a diverse-pose subset; (2) A novel auto-regressive skeleton generation approach featuring efficient joint-based tokenization and hierarchical sequence ordering with randomization strategies; (3) An attention-based architecture for skinning weight prediction incorporating topology-aware joint attention; and (4) A differentiable optimization-based animation method that produces stable, high-quality animation for diverse object categories without requiring extensive computational resources or manual effort.

\section{Related works}
\label{sec:related}

\boldstartspace{Skeleton generation.} Skeleton generation methods for 3D models fall into two main groups. The first leverages templates or additional inputs. Pinocchio \cite{baran2007automatic} pioneered template-fitting for automatic skeleton extraction, while Li et al. \cite{li2021learning} employed deep learning for human joint estimation with a given skeleton template. Some recent works \cite{chu2024humanriglearningautomaticrigging, Guo_2025_CVPR, sun2024drivediffusionbasedriggingempowers} continue this line for humanoid skeleton generation. A significant limitation of these approaches is their inability to generalize effectively to diverse object categories. 
Other methods in this group require additional inputs such as point cloud sequences \cite{xu2022morig}, mesh sequences \cite{de2008automatic, james2005skinning}, manual annotations \cite{mixamo}, or video data \cite{yang2022banmo, wu2023magicpony, song2024reacto, zhang2024magicpose4d, yao2025riggs, song2024moda, sun2024ponymation, li2024fauna}. 
The second group works without templates or annotations.
Traditional approaches \cite{au2008skeleton, cao2010point, huang2013l1, tagliasacchi2012mean, lin2021point2skeleton} extract curve skeletons and often produce overly dense joints unsuitable for animation. 
Modern deep learning methods like Xu et al. \cite{xu2019predicting} and RigNet \cite{xu2020rignet} learn directly from limited datasets containing fewer than 3,000 rigged models. Despite their innovations, these methods depend extensively on carefully crafted features and impose restrictive assumptions regarding shape orientation, substantially constraining their effectiveness when confronted with complex mesh topologies.

With the exponential growth of 3D datasets \cite{deitke2023objaverse, deitke2024objaverse} and the success of auto-regressive approaches in 3D generation \cite{siddiqui2024meshgpt, chen2024meshanything, chen2024meshanythingv2, tang2024edgerunner}, the field has seen significant advances in skeleton generation. MagicArticulate \cite{song2025magicarticulate} pioneered the formulation of skeleton generation as an auto-regressive problem and introduced Articulation-XL, a large-scale 3D dataset with rigging information. Several recent works \cite{liu2025riganything, zhang2025unirig} have also successfully incorporated auto-regressive transformer architectures for skeleton generation, further validating this approach. In our work, we substantially expand the Articulation-XL dataset from 33k to 59.4k rigged models, including a diverse pose subset containing 11.4k examples. We leverage auto-regressive transformers for skeleton generation, introducing two key innovations: an efficient tokenization method for skeletal structures and a hierarchical sequence ordering strategy with randomization that enhances bidirectional learning capabilities. 

\boldstartspace{Skinning weight prediction.}
Following skeleton generation, automatic rigging requires skinning weights prediction to establish joint influence on mesh vertices. Traditional geometric approaches \cite{dionne2013geodesic, jacobson2011bounded, dodik2024robust, baran2007automatic} assign weights based on vertex-joint distances—a method that proves inadequate for complex topologies. Learning-based approaches \cite{liu2019neuroskinning, xu2020rignet, mosella2022skinningnet, 10.1145/3451262} consistently integrate graph neural networks (GNN) with geometric distance cues for skinning weight prediction. However, these GNN-based methodologies face significant limitations in scalability and struggle to generalize effectively across 3D data with diverse spatial orientations. MagicArticulate \cite{song2025magicarticulate} formulates skinning weight prediction as a functional diffusion problem \cite{zhang2024functional}, but suffers from slow inference and limited generalization. We instead introduce an attention-based network that strategically incorporates skeleton graph distances, enabling more robust skinning weight prediction with substantially enhanced generalization across diverse object categories. Concurrent works \cite{zhang2025unirig, deng2025anymate} similarly leverage cross-attention between surface points and bones to learn skinning weights.

\boldstartspace{3D animation.} With the generated rigging, the next step is to animate the 3D models. In contrast to earlier work that focuses on human motion generation \cite{jiang2023motiongpt, tevet2022motionclip, zhang2024motiondiffuse, guo2022generating, zhang2024large, tevet2022human, song20213d, song2023unsupervised, shen2024hmr, shen2025adhmr}, we aim to animate diverse 3D object categories that can be rigged. Our pipeline uses a reference video as motion guidance to animate the rigged mesh. The field of 4D generation has experienced rapid growth recently, spanning text/image-to-4D generation \cite{yin20234dgen, ren2023dreamgaussian4d, ling2024align, bah2024tc4d, miao2024pla4d, yuan20244dynamic, chen2024ct4d, zeng2024trans4d, liang2024diffusion4d, bahmani20244d, sun2024eg4d, lin2024phy124, xuphys4dgen, zhao2023animate124} and video-to-4D generation \cite{gao2024gaussianflow, jiang2023consistent4d, zeng2024stag4d, wu2024sc4d, zhang2024magicpose4d, zhang20244diffusion, li2024dreammesh4d, fu2024sync4d, zhu2025ar4d, yao2025sv4d, xie2024sv4d, chen2025v2m4, ren2024l4gm, wang2024animatabledreamer}. For a comprehensive overview, we refer readers to the survey by Miao et al. \cite{miao2025advances}. However, most existing 4D generation approaches do not take a specific 3D object as input to generate animations targeted for that object.

Many notable attempts have been made to address object animation.  AnyMole \cite{yun2025anymole} introduced a motion in-betweening method for diverse categories with context motions. AnyTop \cite{gat2025anytop} proposed animating 3D rigged meshes using a diffusion model without explicit motion guidance, given only an input skeleton. Millan et al. \cite{millan2025animating} presented related work but focused exclusively on humanoid models with SMPL \cite{SMPL:2015} proxies. Animate3D \cite{jiang2024animate3d} proposed animating 3D objects with a Multi-view Video Diffusion Model, which requires extensive training. MotionDreamer \cite{uzolas2025motiondreamer} attempted to animate 3D models without rigging but produced suboptimal motion quality. The most recent related work, AKD \cite{li2025akd}, takes a 3D model, manually adds a skeleton, and uses \cite{baran2007automatic} to predict skinning weights. They apply animation to this rigged model using video-based score distillation sampling (SDS). However, their approach is computationally intensive (requiring approximately 25 hours per object) and produces unstable animations with noticeable jittering artifacts. In contrast, we propose an optimization-based method that requires no neural network parameters to achieve more stable animations for diverse object categories by combining our generated rigging with reference video guidance.

\section{Automatic rigging} 

Our automatic rigging framework features two sequential modules. 
First, we deploy an auto-regressive transformer to infer a structurally valid skeleton from a raw 3D mesh (\Cref{ar}). 
Subsequently, this skeleton and the original mesh are processed by an attention-based architecture to predict precise per-vertex skinning weights (\Cref{skin}). To facilitate large-scale learning, we introduce \ourdata{} (\Cref{data}), a comprehensive dataset comprising 59.4k 3D models with high-quality rigging.

\subsection{Dataset: Articulation-XL2.0}
\label{data} 
We present \ourdata{}, an expanded version of Articulation-XL proposed in \cite{song2025magicarticulate}. Our dataset incorporates multiple geometric data types from Objaverse-XL \cite{deitke2023objaverse, deitke2024objaverse} previously excluded, while maintaining the same data filtering process. We further improve quality by eliminating unskinned vertices and conducting manual validation, yielding over 48k high-quality rigged 3D models. 

Recognizing that models in our primary dataset are predominantly in rest pose configurations, thus limiting generalization capacity to novel articulations, we have constructed a diverse-pose subset.
By identifying the intersection between high-quality animation data from Diffusion4D \cite{liang2024diffusion4d} and our rigged model corpus, we extract 7.3k deformed meshes with corresponding rigging information from animation frames exhibiting maximal deviation from rest pose configurations. 
To counterbalance the predominance of humanoid morphologies in this subset, we supplement with 4.1k models generated using SMALR \cite{Zuffi:CVPR:2017, Zuffi:CVPR:2018} with parameterizations derived from 41 distinct animal scans and randomized valid poses. 
The resulting 11.4k diverse-pose dataset significantly enhances performance on unseen poses, as validated in our experiments. \textbf{We will release \ourdata{}, a comprehensive collection of 59.4k high-quality rigged models, to facilitate future research.} Dataset statistics and examples are provided in the appendix.

\subsection{Auto-regressive skeleton generation}
\label{ar}

\begin{figure*}
  \centering
\includegraphics[width=\textwidth]
{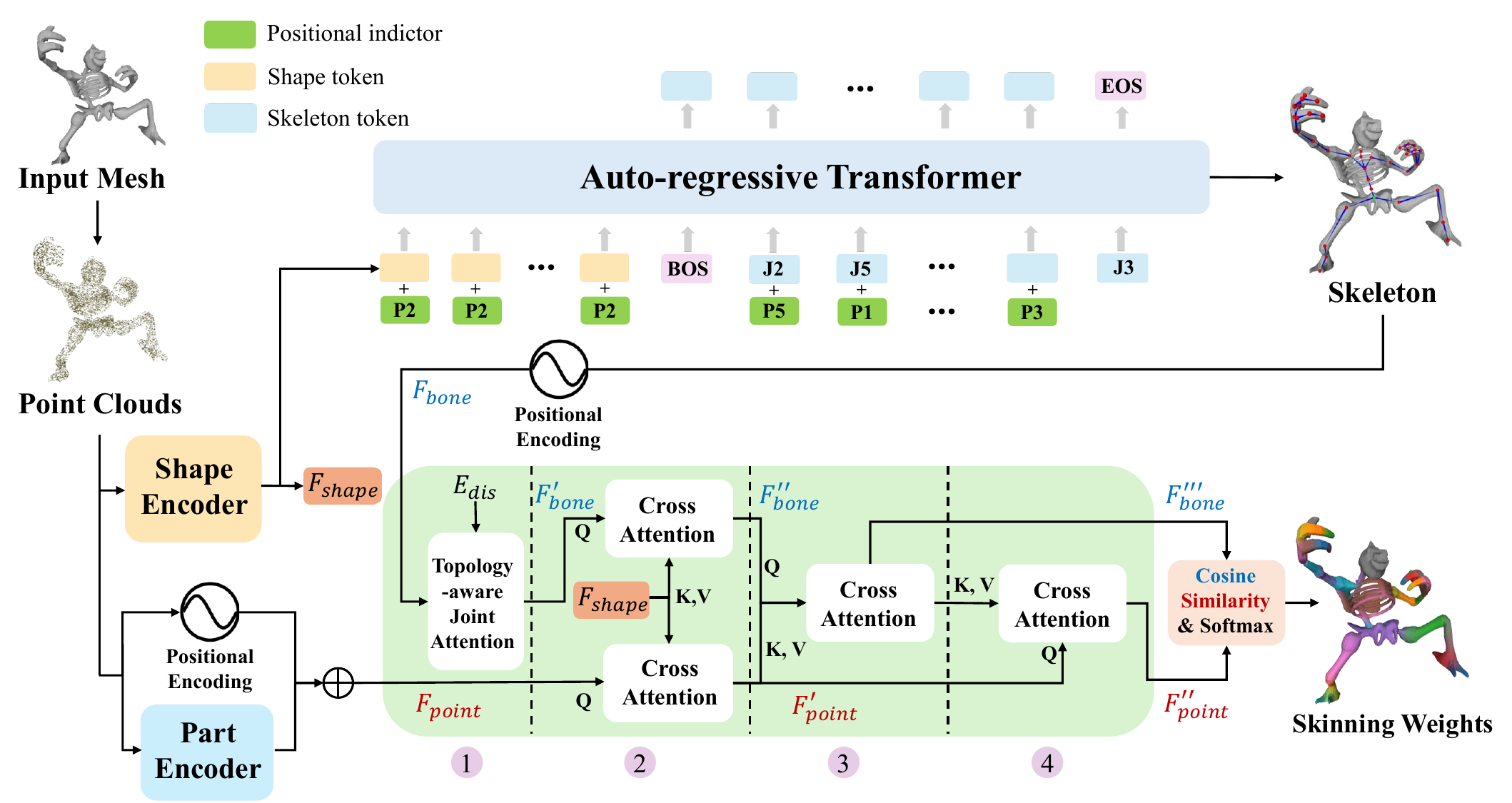}
  \caption{\textbf{Overview of our automatic rigging pipeline.} Given a 3D mesh, we first sample point clouds with normals, then generate a skeleton using an auto-regressive transformer. The point clouds and skeleton are processed through an attention-based network with four key operations: \purplecircle{1} bone feature enhancement via topology-aware joint attention, \purplecircle{2} global context integration through cross-attention with shape latents, \purplecircle{3} bone-point interaction via cross-attention, and \purplecircle{4} point feature refinement. Finally, cosine similarity and softmax normalization produce the skinning weights.}
  \label{method}
  \vspace{-10pt}
\end{figure*}

We formulate skeleton generation as a shape‐conditioned sequence modeling problem. Given an input mesh
$\mathcal{M}$, we employ an auto-regressive framework (\Cref{method} top) to predict a skeleton $\mathcal{S}$ consisting of 3D joint positions $\mathbf{J} \in \mathbb{R}^{j \times 3}$ and topological bone connections $\mathbf{B} \in \mathbb{N}^{b \times 2}$ defined by joint indices.
Our framework consists of three key components: \textit{joint-based skeleton tokenization}, \textit{hierarchical sequence ordering with randomization}, and \textit{shape-conditioned auto-regressive generation}. Together, these components enable accurate, efficient skeleton generation across varied object structures without relying on predefined templates.

\boldstartspace{Joint-based skeleton tokenization.} 
In \cite{song2025magicarticulate}, skeletons are encoded as bone-based sequences: each of the $b$ bones contributes $6$ tokens (the 3D coordinates of its two endpoints), yielding a total sequence length of $6b$ and redundantly repeating joint positions across multiple connected bones. Inspired by \cite{liu2025riganything}, we develop a joint-based tokenization strategy that represents each of the $j$ joints by its 3D coordinates and parent index, producing a sequence of length $4j$. Since a tree-structured skeleton satisfies $j = b + 1$, this yields $4j < 6b$ whenever $j>3$, making the joint-based representation more compact. Unlike \cite{liu2025riganything}, which projects joint positions into high-dimensional feature spaces via MLPs, we discretize normalized joint coordinates into a $128^3$ grid and append the parent index, producing discretized token sequences that serve as input to our auto-regressive transformer. In practice, we assign the root joint a parent index of~$0$ and offset all other parent indices by $+1$ (subtracting~$1$ during detokenization).

\boldstartspace{Sequence ordering.}
While our joint-based tokenization provides a compact representation, the sequential ordering of tokens significantly affects skeletal coherence and model performance. For skeleton modeling with joint positions and parent indices, tokens can be sequenced using either spatial ordering (ascending z-y-x coordinates, as in \cite{song2025magicarticulate}) or hierarchical ordering (breadth-first traversal of the skeletal tree structure).
Our experiments demonstrate that spatial ordering frequently produces disconnected skeletons, as child joints generated before their parents create invalid parent references (see \Cref{ablation} and appendix for comparisons). We therefore adopt hierarchical ordering, applying spatial sorting only among joints at the same hierarchical level.

Additionally, inspired by \cite{yu2024randomized}, we enhance bidirectional learning capability through sequence randomization. We group the $4$ tokens of each joint together and randomly shuffle these groups, incorporating target-aware positional indicators $\mathbf{P} = [\mathbf{p}_0, \mathbf{p}_1, ..., \mathbf{p}_{j-1}]$ to guide the generation process. Specifically, all tokens within a joint group share a positional indicator signaling which joint will be generated next. To identify the first joint group, we additionally incorporate positional indicators into shape tokens $\mathbf{T}_{shape}$ that precede skeleton tokens $\mathbf{T}_{skel}$:

\begin{equation}
    \mathbf{T} = [\mathbf{T}_{shape}, \mathbf{T}_{skel}] + \mathbf{P} = [\mathbf{T}_{shape} + \mathbf{p}_{0}, \mathbf{T}_{skel}^{0} + \mathbf{p}_{1}, ..., \mathbf{T}_{skel}^{j-2} + \mathbf{p}_{j-1}, \mathbf{T}_{skel}^{j-1}]. 
    \label{token}
\end{equation}

\boldstartspace{Shape-conditioned auto-regressive generation.}  With our tokenization strategy and sequence ordering established, we now describe the auto-regressive generation process.
We sample 8,192 points with normals from the input mesh as shape conditioning and encode them using a pre-trained shape encoder \cite{zhao2024michelangelo}. This fixed-length shape token sequence $\mathbf{T}_{shape}$ precedes the transformer's skeleton sequence, with $\textless\!\mathrm{bos}\!\textgreater$
 and $\textless\!\mathrm{eos}\!\textgreater$ tokens marking skeleton boundaries (omitted in \Cref{token}). We adopt OPT-350M \cite{zhang2022opt} as our decoder-only transformer architecture, training with cross-entropy loss for next-token prediction:
\begin{equation} 
    \mathcal{L}_{pred} = \mathrm{CE}(\mathbf{T}, \mathbf{\hat{T}}),
\end{equation}
where $\mathbf{T}$ and $\mathbf{\hat{T}}$ represent ground truth and predicted token sequences.
During inference, generation begins with shape tokens and \textit{sequential} positional indicators, proceeding auto-regressively until producing the $\textless\!\mathrm{eos}\!\textgreater$ token, followed by detokenization to recover the complete skeleton.

\subsection{Attention-based skinning weight prediction}
\label{skin}
\vspace{-5pt}
In this section, we present an attention-based network for predicting per-vertex skinning weights that determine how the mesh deforms in response to skeleton articulation.

\vspace{-5pt}
\boldstartspace{Network architecture.}
The network architecture is illustrated in the bottom of \Cref{method}.
Our pipeline begins by sampling $n$ points with normals from the input mesh. These points are processed through positional encoding and a part encoder from PartField \cite{liu2025partfield} to obtain part-aware point embeddings $\mathbf{F}_{point} \in \mathbb{R}^{n \times d}$ that combine spatial information with part features. We incorporate part-aware features because parts and bones exhibit strong anatomical correspondence, providing valuable structural guidance for skinning weight prediction. In parallel, we construct bone-based coordinates $\in \mathbb{R}^{j \times 6}$ by concatenating each joint’s parent position with its own position—for the root joint, its position is duplicated to fill both coordinate slots. These bone coordinates similarly undergo positional encoding to produce bone embeddings $\mathbf{F}_{bone} \in \mathbb{R}^{j \times d}$. Additionally, we feed the sampled points with normals into a pre-trained shape encoder \cite{zhao2024michelangelo} to extract global shape latents $\mathbf{F}_{shape} \in \mathbb{R}^{257 \times d}$.

The architecture then performs a series of attention operations \cite{vaswani2017attention}:(1) Bone feature enhancement. We first apply self-attention using the topology-aware joint attention on the bone embedding to obtain enhanced bone features $\mathbf{F}_{bone}^{\prime}$. (2) Global context integration. Cross-attention is performed between global shape latents (as context) and both point and bone features, generating updated features $\mathbf{F}_{point}^{\prime}$ and $\mathbf{F}_{bone}^{\prime\prime}$. (3) Bone-point interaction. Cross-attention uses the updated bone features as queries and $\mathbf{F}_{point}^{\prime}$ as keys/values to produce refined bone features $\mathbf{F}_{bone}^{\prime\prime\prime}$. (4) Point feature refinement. Final cross-attention between refined bone features $\mathbf{F}_{bone}^{\prime\prime\prime}$ (as context) and point features $\mathbf{F}_{point}^{\prime}$ produces the final point features $\mathbf{F}_{point}^{\prime\prime}$. Finally, the network computes cosine similarity scores and applies softmax normalization to produce skinning weights:
\begin{equation}
    \mathbf{W} = \text{softmax}\left(\alpha\frac{\mathbf{F}_{point}^{\prime\prime}\mathbf{F}_{bone}^{\prime\prime\prime\top}}{\left\|\mathbf{F}_{point}^{\prime\prime}\right\|\left\|\mathbf{F}_{bone}^{\prime\prime\prime}\right\|}\right). 
\end{equation}
where $\alpha$ is a learnable scaling parameter. We optimize the network using cross-entropy loss during training.

\vspace{-5pt}
\boldstartspace{Topology-aware joint attention.} 
While the basic architecture provides effective weight prediction, our experiments revealed that explicitly modeling skeletal structure significantly enhances performance. Our ablation studies demonstrate that using bone-based coordinates $\in \mathbb{R}^{j \times 6}$ rather than joint coordinates $\in \mathbb{R}^{j \times 3}$ substantially improves performance (see \Cref{ablation}), highlighting the importance of inter-joint relationships within the skeletal structure. 

To further leverage topological structure, we propose Topology-aware Joint Attention (TAJA), which augments standard self-attention with relative positional encodings derived from skeletal graph distances.
To implement TAJA, we first compute a graph distance matrix $\mathbf{D} \in \mathbb{R}^{j \times j}$ from the skeletal structure, then transform these distances into continuous embeddings through quantization and projection operations, yielding position embeddings $\mathbf{E}{_{dis}} \in \mathbb{R}^{j \times j \times h}$, where $h$ is the number of attention heads. The attention mechanism is then modified as:
\begin{equation}
\text{Attention}(\mathbf{Q}, \mathbf{K}, \mathbf{V}, \mathbf{E}_{{dis}}) = \text{softmax}\left(\frac{\mathbf{Q}\mathbf{K}^T}{\sqrt{d_k}} + \lambda\mathbf{E}_{{dis}}\right)\mathbf{V},
\end{equation}
where $\lambda$ is a learnable scaling parameter. This approach explicitly incorporates inter-joint topological relationships, improving the network's capacity to understand skeletal structure and generate more accurate skinning weights.

\section{Video-guided 3D animation}
\vspace{-5pt}
With the generated skeleton and skinning weights, we transform static meshes into animation-ready assets. This section presents our optimization-based approach for automatically animating rigged 3D models with video guidance.

\vspace{-5pt}
\boldstartspace{Animation pipeline.} 
Our animation process begins by rendering the rigged mesh as the initial frame $\mathbf{I}_{0}$. Using this as a conditioning image, we leverage recent text-to-video generation models \cite{jimengai2025, klingai} that can maintain object identity while creating plausible motion sequences. With a text prompt describing the desired animation, these models generate a video sequence $V = \{\mathbf{I}_{0}, \mathbf{I}_{1}, ..., \mathbf{I}_{n-1}\}$ comprising $n$ frames. Given this reference video sequence $V$, we jointly optimize per-frame joint rotations and global root motion of the 3D mesh to align the resulting animation with the generated video sequence.

\vspace{-5pt}
\boldstartspace{Differentiable optimization framework.}
For each frame $i \in \{1, 2, ..., n-1\}$ excluding the first frame, we optimize both root motion parameters $(\mathbf{Q}_{root}^{i}, \mathbf{T}_{root}^{i})$ and joint-specific rotations $Q_{joint}^{i} = \{\mathbf{Q}_{0}^{i}, \mathbf{Q}_{1}^{i}, ..., \mathbf{Q}_{j-1}^{i}\}$, where $\mathbf{Q} \in \mathbb{R}^{4}$ represents rotation as a unit quaternion and $\mathbf{T} \in \mathbb{R}^{3}$ denotes translation. For the first frame (rest pose), we initialize all transformations with identity quaternions and zero translations, which remain fixed during optimization. All subsequent frames are similarly initialized before optimization begins. Our optimization process incorporates rendering losses, tracking losses, and regularization terms:
\begin{equation}
    \mathcal{L} = \underbrace{(\mathcal{L}_{rgb} + \mathcal{L}_{mask} + \mathcal{L}_{flow} + \mathcal{L}_{depth})}_{\text{rendering losses}} + \underbrace{(\mathcal{L}_{joint\_track} + \mathcal{L}_{vertex\_track})}_{\text{tracking losses}} + \mathcal{L}_{reg}.
\end{equation}
For rendering losses, we utilize differentiable rendering via Pytorch3D \cite{ravi2020pytorch3d} to generate predicted frames $\mathbf{I}^{\prime}_{i}$ and compute RGB, mask, optical flow, and depth discrepancies between these predictions and the corresponding reference video frames. The optical flow and depth for video frames are extracted using off-the-shelf methods \cite{video_depth_anything, Morimitsu2025DPFlow}. The tracking losses incorporate a 2D joint tracking term and a 2D vertex tracking term that leverage Cotracker3 \cite{karaev2024cotracker3} to trace selected points throughout the video sequence. We project our optimized 3D joints and deformed mesh vertices into 2D space and minimize their distance to the corresponding tracked 2D keypoints. To address occlusion challenges, we implement visibility detection mechanisms for both joints and vertices. For joints, we define visibility based on ray-mesh intersection: \textit{a joint is considered visible if the ray projected from the camera to the joint intersects the mesh surface exactly once}. We employ the $\text{ray\_mesh\_intersect}$ function from libigl \cite{libigl} to compute these joint visibility masks. For vertex visibility, we leverage the rasterization output from Pytorch3D to determine visible surface points. These visibility masks, derived from the first frame, ensure that our tracking losses are applied consistently throughout the sequence based on initial visibility, preventing optimization artifacts from elements that are occluded in the reference pose. We further incorporate regularization terms that enforce frame-to-frame motion smoothness. Complete mathematical formulations of all loss components are provided in the appendix.

\vspace{-5pt}
\section{Experiments}
\label{sec:exp}
\vspace{-5pt}
\subsection{Experimental setup}
\vspace{-5pt}
\boldstartspace{Datasets.} 
We train our models on the \ourdata{} dataset introduced in \Cref{data}, which contains over 48k high-quality samples from Objaverse-XL \cite{deitke2023objaverse, deitke2024objaverse} as the main subset and 11.4k samples from the diverse-pose subset. For model training, we utilize over 46k samples from the main subset and 10.9k from the diverse-pose subset.
For evaluation,  we employ three distinct test sets: \ourdata{}-test (2k data from the main set), \res{}-test \cite{ModelsResource2019, xu2020rignet} (270 upright, front-facing models with no overlap with \ourdata{}, enabling assessment of cross-dataset generalization), and a 500-mesh portion of the diverse-pose subset specifically selected to evaluate model performance under varied poses.

 \vspace{-5pt}
\boldstartspace{Implementation details.} 
 To enhance robustness and generalization capabilities, we apply geometric data augmentations (scaling, shifting, rotation transformations) and pose augmentation—articulating the training samples with their ground truth skeleton and skinning weights to simulate diverse poses. Further implementation details are provided in the appendix.

\vspace{-5pt}
\subsection{Skeleton generation results}
\vspace{-5pt}
\boldstartspace{Baselines and metrics.}
We include four comparison methods as baselines: Pinocchio \cite{baran2007automatic}, which fits predefined skeleton templates to input meshes. RigNet \cite{xu2020rignet}, a learning-based model that employs graph convolutions to infer joint locations. MagicArticulate \cite{song2025magicarticulate}, an auto-regressive framework for skeleton generation, and the concurrent method UniRig \cite{zhang2025unirig}, which similarly uses an auto-regressive transformer approach. All methods are evaluated on \ourdata{} and \res{} test sets, as well as our diverse-pose subset. We evaluate skeleton generation quality using three Chamfer Distance–based metrics from \cite{xu2019predicting, xu2020rignet}: CD-J2J (joint-to-joint), CD-J2B (joint-to-bone) and CD-B2B (bone-to-bone). These metrics measure the spatial alignment between generated and ground truth skeletons, where lower values indicate better performance.

\begin{table*}[t]
  \centering
  \caption{\textbf{Quantitative comparison of skeleton generation.} We evaluate each method on three benchmarks using CD-J2J, CD-J2B, and CD-B2B—all reported in units of $10^{-2}$. Lower values indicate better alignment. * denotes models trained on \ourdata{} including the diverse-pose subset; unmarked models were trained without it. Bold and underlined numbers denote the best and second-best results, respectively.
  }
  \vspace{-5pt}
  \label{com_skel}
  \begin{tabular}{cccccccccc}
    \toprule
    \multirow{2}{*}{Method} & \multicolumn{3}{c}{\ourdata{}} &  \multicolumn{3}{c}{\res{}} & \multicolumn{3}{c}{Diverse-pose} \\ \cmidrule{2-10}
  &  J2J $\downarrow$ & J2B $\downarrow$& B2B $\downarrow$&J2J $\downarrow$& J2B $\downarrow$& B2B $\downarrow$&J2J $\downarrow$& J2B $\downarrow$& B2B $\downarrow$  \\
    \midrule
     Pinocchio  &8.324 & 6.612 & 5.485 & 6.852 & 4.824 & 4.089  & 7.967 & 6.411 & 5.149 \\
    RigNet &  7.618 & 6.076 & 5.279 & 7.223 & 5.987 & 4.329 & 7.751 & 6.392 & 5.713 \\
      MagicArti. & 3.264 &
2.503 & 2.123 & 4.114 & 3.137
& 2.693  & 4.376 &3.456 & 2.955 \\
UniRig & 3.305	& 2.611	& 2.180	& 3.964	&3.021 &	2.570 & 3.252& 2.569 & 2.077 \\
      Ours   &  \textbf{3.033} &
 \textbf{2.300} &
\textbf{1.923} & \underline{3.841} & \underline{2.881} & \underline{2.475} &  \underline{3.212} & \underline{2.542} & \underline{2.027}   \\
      Ours*   & \underline{3.109} &
\underline{2.370} & \underline{1.983} & \textbf{3.766} & \textbf{2.804} &
\textbf{2.405} & \textbf{2.514} & \textbf{1.986} &
\textbf{1.598}   \\
    \bottomrule
  \end{tabular}
  \vspace{-5pt}
\end{table*}

\begin{figure*}
    \centering
\includegraphics[width=0.85\linewidth]
{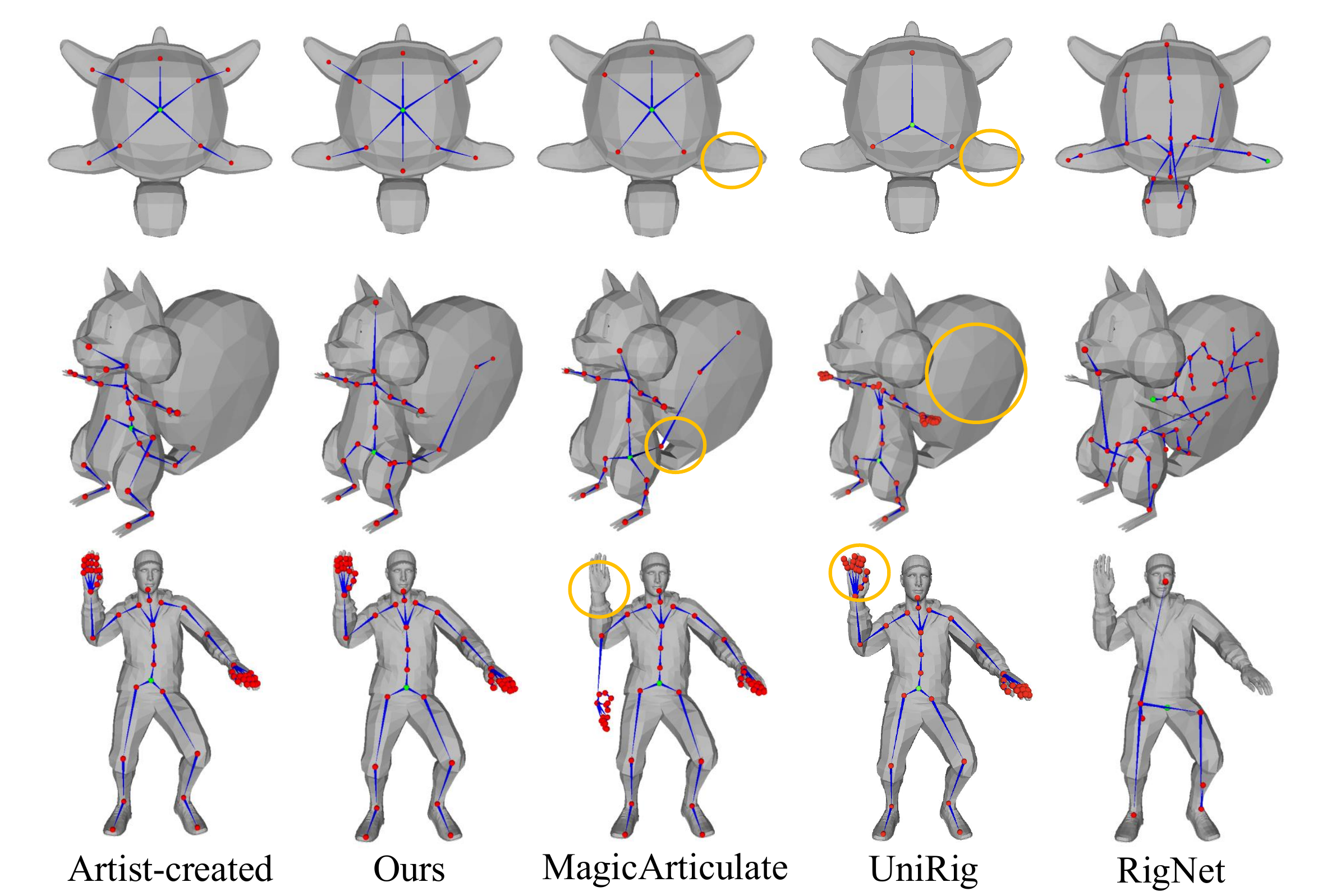}

    \caption{\textbf{Qualitative skeleton generation results}. The data is from \ourdata{}, \res{}, and the diverse-pose subset from top to bottom.} 
    \label{compare_skel}
  \end{figure*}

\vspace{-5pt}
\boldstartspace{Comparison results.} Qualitative results are shown in \Cref{compare_skel} for all three benchmarks. RigNet consistently produces invalid skeletons—its graph-convolutional model fails to converge well when trained on our large‐scale dataset with highly varied orientations. UniRig presents missing and misaligned skeletons, such as missing bones on the turtle limbs and squirrel tail and misaligned skeletons on human hands, as marked in yellow circles. MagicArticulate matches reference skeletons closely on \ourdata{} and \res{}, but exhibits errors in fine details (e.g., missing bones in turtle limbs, incorrect squirrel tail–body junctions) and degrades on the diverse-pose subset, since it was trained only on predominantly rest-pose data without pose augmentation. In contrast, our method yields accurate, structurally correct skeletons across three benchmarks. Importantly, our generated skeletons can even correct omissions in artist-created skeletons, such as a missing turtle head–body connection. \Cref{com_skel} reports quantitative metrics, where we consistently outperform all baselines on every dataset and metric. Notably, incorporating the diverse-pose subset during training leads to marked improvements on the diverse-pose benchmark.

\begin{figure*}
    \centering
\includegraphics[width=\textwidth]
{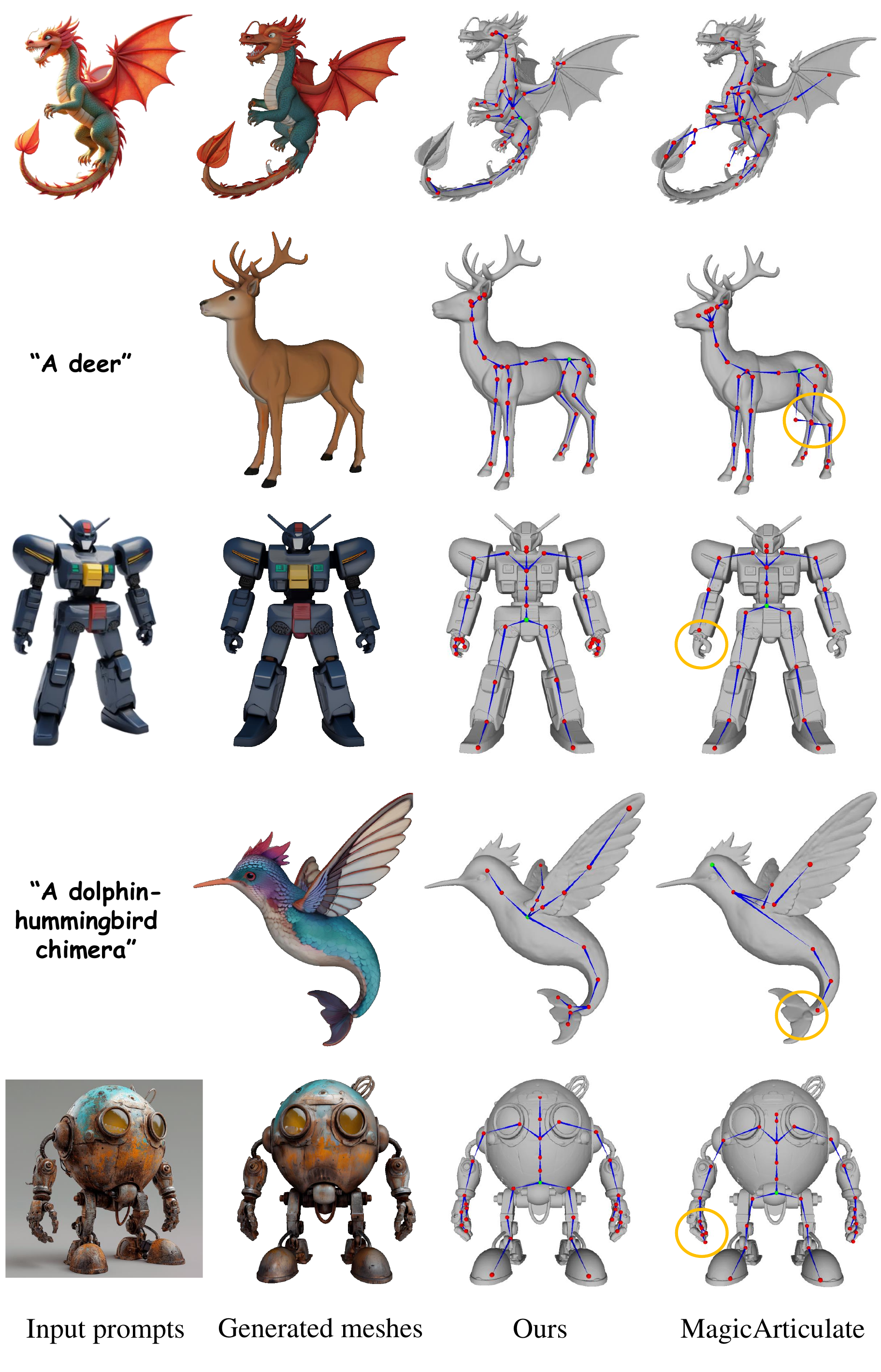}

    \caption{\textbf{Comparison of skeleton results on generated meshes.} The meshes are generated by Tripo 2.0 \cite{tripo3d} and Hunyuan3D 2.0 \cite{hunyuan3d22025tencent}.} 
    \label{supp_skel_ood}
  \end{figure*}

\boldstartspace{Qualitative results on AI-generated meshes.} We evaluate our method's generalization capability on AI-generated meshes from Tripo2.0 \cite{tripo3d} and Hunyuan3D 2.0 \cite{hunyuan3d22025tencent}. As shown in \Cref{supp_skel_ood}, 
we compare our method with MagicArticulate \cite{song2025magicarticulate}. MagicArticulate loses fine details (e.g., the robot’s hand in rows 3 and 5, the dolphin-hummingbird chimera’s tail and wings in row 4, marked in yellow) and produces misaligned skeletons (dragon’s tail in row 1, deer’s legs in row 2). By contrast, our approach consistently generates valid, robust skeletons across all categories.

\vspace{-5pt}
\subsection{Skinning weight prediction results}
\vspace{-5pt}
\boldstartspace{Baselines and metrics.} 
We compare our method for skinning weight prediction against three baselines: Geodesic Voxel Binding (GVB) \cite{dionne2013geodesic}, a geometry-based technique available in Autodesk Maya \cite{AutodeskMaya2024}, RigNet \cite{xu2020rignet}, and MagicArticulate \cite{song2025magicarticulate}. We also evaluate these three methods on \ourdata{} and \res{} test sets, as well as our diverse-pose subset. Skinning weight quality is evaluated using three metrics: precision, recall, and L1-norm error. Precision is the fraction of predicted weights $>1\mathrm{e}{-4}$ that are correct, and recall is the fraction of true weights $>1\mathrm{e}{-4}$ we recover. The L1-norm error reports the average absolute deviation between predicted and ground truth weights over all vertices. Deformation error results are provided in the appendix.

\begin{figure*}
  \centering
\includegraphics[width=\textwidth]
{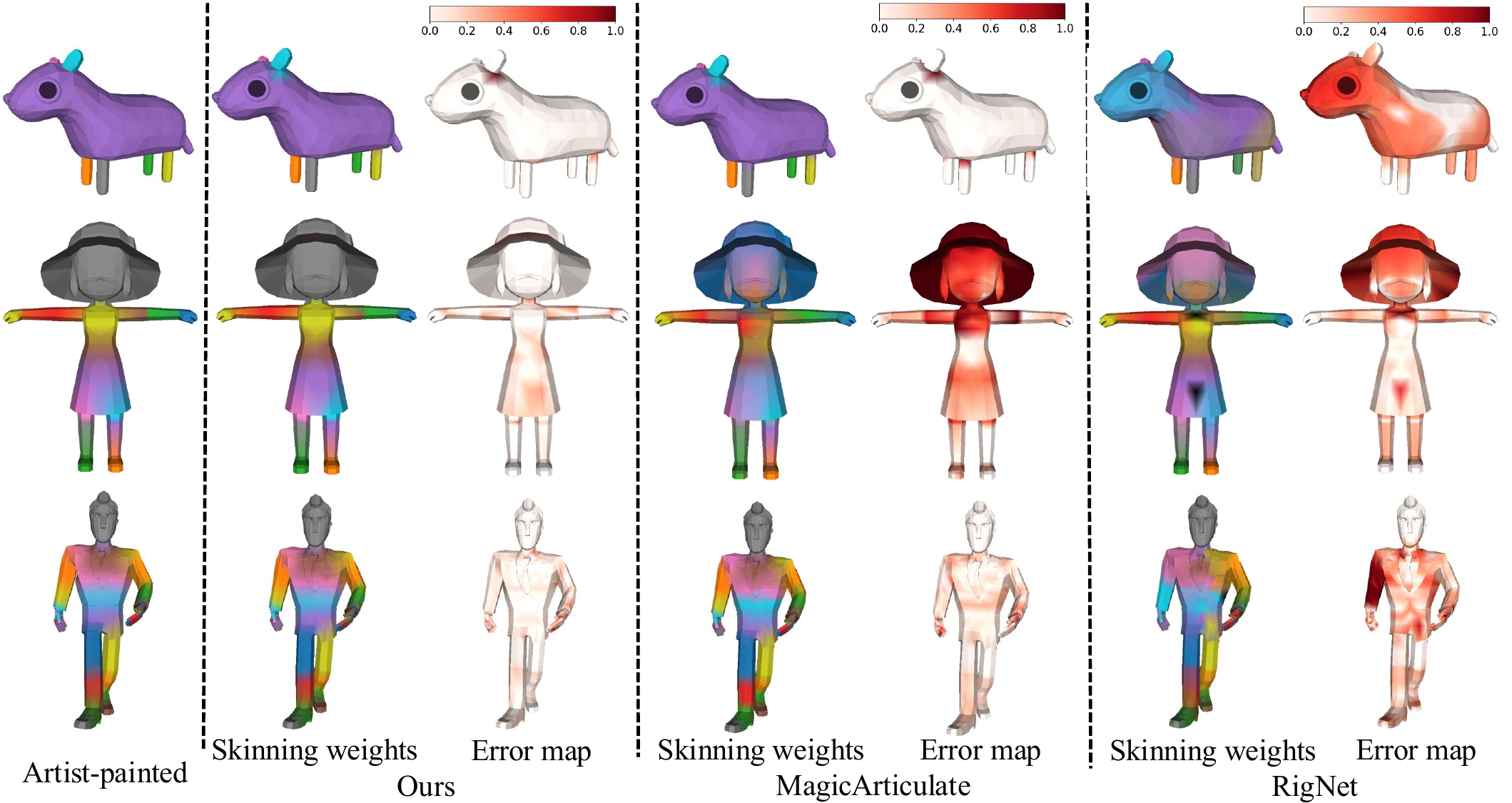}
  \vspace{-5pt}
  \caption{\textbf{Qualitative skinning weight prediction results}. The data is from \ourdata{}, \res{}, and the diverse-pose subset from top to bottom. Each example shows the predicted weight visualization alongside its L1 error map. Additional results are provided in the appendix.
  }
  \label{compare_skin}
  \vspace{-3pt}
\end{figure*}

\vspace{-5pt}
\boldstartspace{Comparison results.}
\Cref{compare_skin} visualizes each method’s predicted skinning weights alongside their L1 error maps. Our method produces more accurate weight distributions with substantially lower errors across all benchmarks. RigNet exhibits large errors on all examples, while MagicArticulate’s functional diffusion performs well on \ourdata{} and the diverse-pose subset but degrades on \res{}, revealing limited cross-dataset generalization. Quantitative results in \Cref{com_skin} confirm these observations, with our method outperforming all baselines on every metric and dataset. Moreover, our approach runs faster—achieving per-example inference speeds that are 1.75$\times$, 45$\times$, and 59$\times$ those of RigNet, MagicArticulate, and GVB, respectively (see appendix for details).

\begin{table*}[t]
  \centering
  \caption{\textbf{Quantitative comparison of skinning weight prediction.} We evaluate our approach against GVB, RigNet, and MagicArticulate. For Precision (Prec.) and Recall (Rec.), higher values indicate better accuracy and coverage. For average L1-norm error (L1), lower is better. Here, * denotes models trained on \ourdata{} with the diverse-pose subset.}
  \label{com_skin}
  \begin{tabular}{cccccccccc}
    \toprule
    \multirow{2}{*}{Method} & \multicolumn{3}{c}{\ourdata{}} &  \multicolumn{3}{c}{\res{}} & \multicolumn{3}{c}{Diverse-pose} \\ \cmidrule{2-10}
  &  Prec. $\uparrow$ & Rec. $\uparrow$ &  L1 $\downarrow$&Prec. $\uparrow$ & Rec. $\uparrow$&  L1 $\downarrow$&Prec. $\uparrow$  & Rec. $\uparrow$&  L1 $\downarrow$  \\
    \midrule
     GVB  & 72.9\% &  65.5\% & 0.745 & 69.3\% & 79.2\% & 0.687
  & 75.2\% & 64.9\% & 0.786 \\
    RigNet &  73.7\% & 66.1\% & 0.729 & 65.7\% & 80.2\% & 0.707 & 74.7\% &  65.4\% & 0.746 \\
      MagicArti. & 74.6\% & 71.3\% & 0.451 &  68.1\% & 80.7\%
 & 0.642 & 74.9\% & 68.4\% & 0.479\\
      Ours   &  \underline{87.6\%} & \textbf{74.0\%} & \underline{0.335} & \underline{79.7\%} & \textbf{81.6\%} & \underline{0.443}   & \underline{83.6\%} & \underline{72.2\%} & \underline{0.405}   \\
      Ours*   &  \textbf{87.9\%} & \underline{73.8\%} & \textbf{0.333}& \textbf{79.8\%} & \underline{81.5\%} & \textbf{0.442}   & \textbf{86.4\%} & \textbf{72.8\%} & \textbf{0.353}    \\
    \bottomrule
  \end{tabular}
  \vspace{-5pt}
\end{table*}

\vspace{-5pt}
\subsection{3D animation results}
\begin{figure*}
  \centering
  \includegraphics[width=\textwidth]
{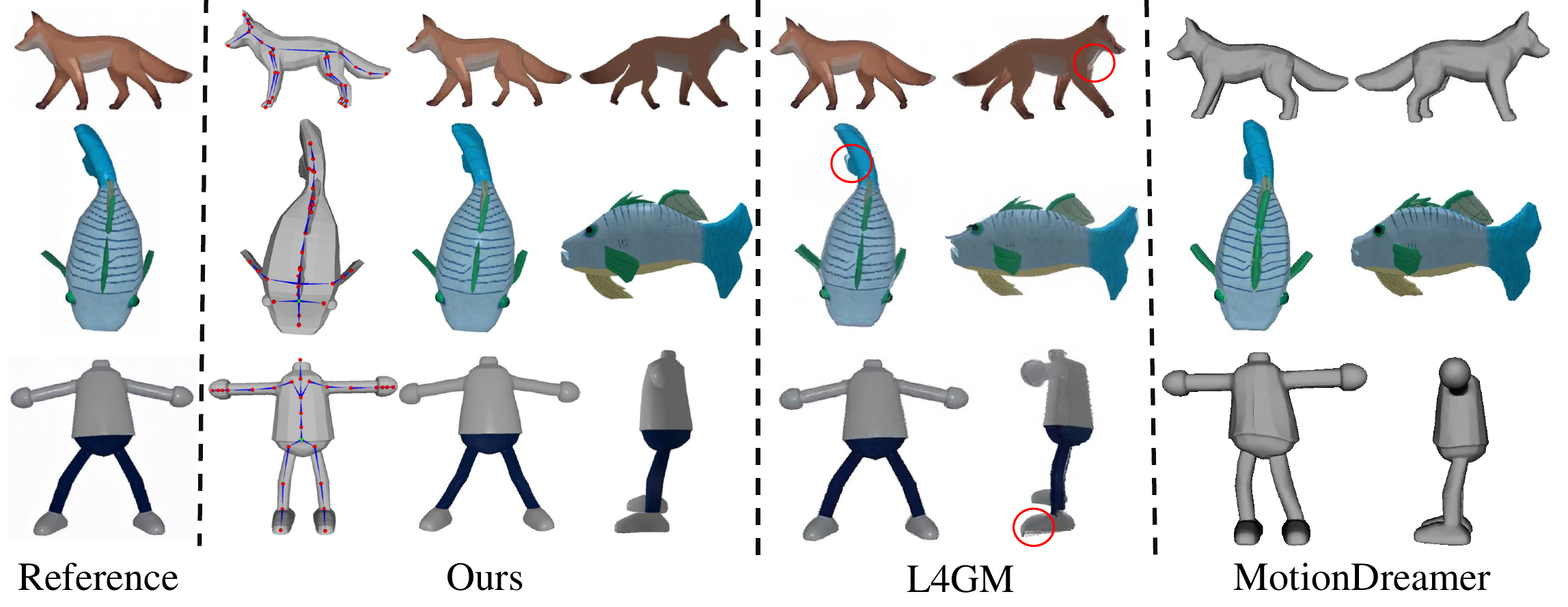}
  \vspace{-8pt}
  \caption{\textbf{Comparison of animation results.} We present our generated skeletons and corresponding video‐guided animations. \textbf{The shapes with skeletons represent rest poses.} While L4GM \cite{ren2024l4gm} aligns its reference views closely with the input video, it consistently exhibits distortions (highlighted in red). MotionDreamer’s \cite{uzolas2025motiondreamer} animations are subtle and can introduce unintended deformations in rigid parts (e.g., the humanoid torso). In contrast, our method delivers accurate, artifact‐free animations using fully generated rigging. Videos are included in the project page.}
  \label{compare_anime}
  \vspace{-5pt}
\end{figure*}

\vspace{-5pt}
\boldstartspace{Baselines.} We compare our animation results with L4GM \cite{ren2024l4gm} for video-to-4D generation and MotionDreamer \cite{uzolas2025motiondreamer} for 3D mesh animation. To ensure a fair evaluation, L4GM is given the same input videos and its multi-view synthesis for the first frame is replaced with ground-truth renderings of the input 3D model. MotionDreamer receives the input 3D model along with the same text prompts used for video generation. In \Cref{compare_anime}, some of its outputs appear untextured because its watertight mesh conversion breaks the UV mappings.

\vspace{-5pt}
\boldstartspace{Comparison results.} As shown in \Cref{compare_anime}, we present our generated skeletons and the corresponding video-guided animations. The shapes with skeletons represent the rest poses.
Although L4GM’s reference views are well aligned with the source video, it repeatedly produces geometric distortions (red highlights), even when provided with ground truth multi-view renderings. MotionDreamer’s animations are subtle and can introduce unintended deformations in rigid parts (e.g., the humanoid torso). By contrast, our approach produces accurate, artifact-free animations using fully generated rigging.

\vspace{-5pt}
\subsection{Ablation studies}
\vspace{-5pt}
\label{ablation}
In this section, we present ablation studies on both skeleton generation and skinning weight prediction. All models are trained on \ourdata{} without the diverse‐pose subset.

\vspace{-5pt}
\boldstartspace{Ablation studies on skeleton generation.} 
We ablate four components—pose augmentation, order randomization, tokenization scheme, and skeleton ordering strategy—to measure their effects on skeleton generation (see \Cref{ab_skel}). Removing pose augmentation degrades performance across all benchmarks, especially on the diverse-pose test. Disabling order randomization similarly reduces accuracy. Bone-based tokenization matches our method’s quality but requires 12 extra training hours and is 1.6$\times$ slower at inference. Finally, replacing hierarchical ordering with spatial ordering preserves CD-J2J and CD-J2B but markedly increases CD-B2B error and often produces disconnected skeletons; see the appendix for visualization comparisons.

\begin{table*}
  \centering
  \caption{\textbf{Ablation studies on skeleton generation.}}
  \vspace{-5pt}
  \label{ab_skel}
  \begin{tabular}{cccccccccc}
    \toprule
    \multirow{2}{*}{Method} & \multicolumn{3}{c}{\ourdata{}} &  \multicolumn{3}{c}{\res{}} & \multicolumn{3}{c}{Diverse-pose} \\ \cmidrule{2-10}
  &  J2J $\downarrow$ & J2B $\downarrow$& B2B $\downarrow$&J2J $\downarrow$& J2B $\downarrow$& B2B $\downarrow$&J2J $\downarrow$& J2B $\downarrow$& B2B $\downarrow$  \\
    \midrule
     w/o pose aug.  & 3.131 &2.451 & 2.223
 & 3.994 & 3.141 & 2.843 & 4.886 & 4.029 & 3.629 \\
 w/o random   &3.166 & 2.431& 2.057 & 3.902 & 3.006 & 2.695 & 3.356 &
2.631 & 2.201    \\
    Bone token &  \underline{3.014} & 2.309 & \underline{1.939} & \underline{3.865}
& \underline{2.940} & \underline{2.524} & 3.269 & \textbf{2.518} &
\underline{2.087} \\
      Spatial order & \textbf{2.982} & \textbf{2.298} & 2.068 & 3.868 & 2.961 & 2.641 & \textbf{3.210} &
2.570 & 2.295 \\
      Ours   &  3.033 &
 \underline{2.300} &
\textbf{1.923} & \textbf{3.841} & \textbf{2.881} & \textbf{2.475} &  \underline{3.212} & \underline{2.542} & \textbf{2.027}  \\
    \bottomrule
  \end{tabular}
  \vspace{-8pt}
\end{table*}

\vspace{-5pt}
\begin{table*}[htbp]
  \centering
  \caption{\textbf{Ablation studies on skinning weight prediction.}}
  \vspace{-5pt}
  \label{ab_skin}
  \begin{tabular}{cccccccccc}
    \toprule
    \multirow{2}{*}{Method} & \multicolumn{3}{c}{\ourdata{}} &  \multicolumn{3}{c}{\res{}} & \multicolumn{3}{c}{Diverse-pose} \\ \cmidrule{2-10}
 &  Prec. $\uparrow$ & Rec. $\uparrow$ &  L1 $\downarrow$&Prec. $\uparrow$ & Rec. $\uparrow$&  L1 $\downarrow$&Prec. $\uparrow$  & Rec. $\uparrow$&  L1 $\downarrow$  \\
    \midrule
Joint embed & 87.1\% &
73.8\% & 0.346 & 79.2\% & 80.8\% & 
0.458 & 82.8\% & \underline{72.2\%} & 0.427\\  
w/o TAJA & 86.6\% & \textbf{74.2\%}
 & 0.348 & 79.2\% & \underline{81.4\%} & 0.450  & 82.8\% & \textbf{72.5\%} & \underline{0.414} \\
w/o part feat. & 87.4\% &
73.8\% & 0.338 & 79.1\% & 81.0\% & 
0.451 & \underline{83.2\%} & 72.1\% & \underline{0.414} \\  
     w/o pose aug. & \textbf{88.0\%} & 73.3\% &
\underline{0.337} & \underline{79.3\%} & 80.7\% & \underline{0.449} & 82.8\% & 70.1\% & 0.444 \\
      Ours   &  \underline{87.6\%} & \underline{74.0\%} & \textbf{0.335} & \textbf{79.7\%} & \textbf{81.6\%} & \textbf{0.443}   & \textbf{83.6\%} & \underline{72.2\%} & \textbf{0.405}    \\
    \bottomrule
  \end{tabular}
  \vspace{-8pt}
\end{table*}

\boldstartspace{Ablation studies on skinning weight prediction.} We evaluate four key components of our skinning weight prediction framework (see \Cref{ab_skin}). First, replacing bone embeddings with joint embeddings increases the average L1-norm error by 4.0\% across all three benchmarks, demonstrating the importance of explicitly modeling bone information. Second, replacing Topology-aware Joint Attention (TAJA) with standard self-attention leads to performance degradation across all benchmarks, highlighting the value of modeling topological relationships between joints. Third, removing part-aware features results in consistent performance drops, confirming their contribution to accurate weight prediction. Finally, eliminating pose augmentation during training increases the L1-norm error on the diverse-pose subset by 9.6\%, demonstrating that pose variation is essential for generalization to novel poses. These findings confirm that each component is crucial to our model's overall accuracy.

\section{Conclusion}
\vspace{-5pt}
In this work, we introduce \ours{}, a unified rigging-and-animation pipeline built on a dataset with 59.4k high-quality rigged models. \ours{} first generates skeletons with an autoregressive transformer that uses joint-based tokenization and hierarchical ordering with randomization to capture skeletal structures. An attention-based network with topology-aware features then predicts skinning weights, followed by an efficient optimization module that produces stable, high-quality animations at low computational cost. Across multiple benchmarks, \ours{} outperforms state-of-the-art methods in skeleton fidelity, skinning accuracy, and animation smoothness.

\section*{Acknowledgements} 
This research is supported by the MoE AcRF Tier 2 grant (MOE-T2EP20223-0001) and the MoE AcRF Tier 1 grant (RG14/22).

\newpage

\appendix

\renewcommand{\thetable}{S\arabic{table}}
\renewcommand{\thefigure}{S\arabic{figure}}

\section*{Appendix}
In this appendix, we provide additional details and experimental results for the main paper, including:

\begin{itemize}
    \item Further details of \ours{} (\Cref{method_detail}) and \ourdata{} (\Cref{detail_data});
    \item Additional experimental results on skeleton generation, skinning weight prediction, and animation (\Cref{additioanl_results});
    \item A discussion of the limitations of our work and future works (\Cref{limit}), and broader impact considerations (\Cref{impact}).
\end{itemize}

\section{More details of \ours{}}
\label{method_detail}

\subsection{Implementation details}
\boldstartspace{Skeleton generation.}
Our skeleton generation begins by encoding each mesh with a pre-trained shape encoder \cite{zhao2024michelangelo}. We first compute its signed distance function using \cite{wang2022dual}, reconstruct a coarse mesh using Marching Cubes \cite{lorensen1998marching}, and then sample 8,192 surface points (with normals). These points are finally encoded into a fixed sequence of 257 shape tokens. 
We normalize input points to $[-0.5, 0.5]$ and apply the same scale and translation to joint positions for alignment. Joint coordinates are then discretized into a $128^{3}$ grids, and parent indices are appended—yielding a token sequence of length $4j$.

During training, we apply pose augmentation with a probability of 0.5. When pose augmentation is applied, each joint has a 0.3 probability of rotation, with rotation angles constrained to the range of $[-60^{\circ}, 60^{\circ}]$. Sequence ordering randomization is annealed following \cite{yu2024randomized}, with a permutation probability $r$ that starts at 1 and falls to 0 (reverting to hierarchical order) over training:

\begin{equation}
    r = \left\{ 
    \begin{aligned}
& 1.0,   \quad & {epoch} \in & [0, E/2], \\
& 1 - \frac{{epoch} - E/2}{E/4}, \quad & {epoch} \in & [E/2, 3E/4], \\
& 0.0,  \quad &  {epoch} \in & [3E/4, E], \\
    \end{aligned}
    \right.
\end{equation}

where ${epoch}$ is the current epoch and $E$ the total number of epochs. Besides, we apply target‐aware positional indicators that mark which joint group the model should generate next. Both shape and skeleton token sequence (except for the final joint group) are augmented with an indicator denoting their next group; the indicator on the shape tokens specifically identifies the initial joint group to generate. The auto-regressive transformer is trained on 8 NVIDIA A100 GPUs (batch size 64 per GPU, effective batch 512) for approximately 3 days and 20 hours.

\boldstartspace{Skinning weight prediction.} During training, we condition the attention-based network for skinning weight prediction on the ground truth skeleton and supervise it with the corresponding skinning weights.
We sample 8,192 surface points (with normals) from each mesh—matching our skeleton pipeline—and assign each point the weights of its nearest vertex. During inference, these predicted weights are transferred back to mesh vertices through nearest-neighbor mapping. 

Training is performed on \ourdata{} with 8 NVIDIA A100 GPUs for roughly 1 day and 6 hours, with a batch size of 16 per GPU. A valid-joint mask—supporting up to 70 joints—adapts the network to the varying skeleton sizes encountered during both training and evaluation.

\boldstartspace{Video-guided 3D animation.} Here, we describe our video-guided 3D animation process in detail. For rendering supervision, we employ four distinct loss functions.  Given a generated video $V = \{\mathbf{I}_{0}, \mathbf{I}_{1}, ..., \mathbf{I}_{n-1}\}$ and the rendered images $\mathbf{I}_{i}^{\prime}$ using Pytorch3D \cite{ravi2020pytorch3d}, we calculate the following rendering losses: 

\begin{equation}
\begin{aligned}
    &\mathcal{L}_{rgb}  = \sum_{i}{\left\|\mathbf{M}_{i} \odot (\mathbf{I}_{i} - \mathbf{I}_{i}^{\prime})\right\|^{2}},
&& \mathcal{L}_{mask} = \sum_{i}{\mathrm{BCE}(\mathbf{M}_{i}, \mathbf{M}_{i}^{\prime})}, \\
   & \mathcal{L}_{depth}  = \sum_{i}{\left\|\mathbf{M}_{i} \odot (\mathbf{D}_{i} - \mathbf{D}^{\prime}_{i})\right\|^{2}},
&& \mathcal{L}_{flow} = \sum_{i}{\left\|\mathbf{M}_{i} \odot (\mathbf{F}_{i} - \mathbf{F}_{i}^{\prime})\right\|^{2}},
    \end{aligned}
\end{equation}

where $\mathbf{M}_{i}$ represents the binary mask of foreground objects in the video frames, and $\mathbf{M}_{i}^{\prime}$ denotes the mask of the 3D object rendered via Pytorch3D. We extract depth maps $\mathbf{D}_{i}$ using the method proposed in \cite{video_depth_anything}, while $\mathbf{D}_{i}^{\prime}$ is obtained directly from the Pytorch3D renderer. We apply scale-shift alignment to handle the scale ambiguity between relative depth $\mathbf{D}_{i}$ and metric depth $\mathbf{D}_{i}^{\prime}$. For optical flow estimation, we compute $\mathbf{F}_{i}$ using the approach from \cite{Morimitsu2025DPFlow}, and derive $\mathbf{F}_{i}^{\prime}$ by projecting the 3D vertex flow onto the 2D image plane. The element-wise multiplication operator $\odot$ indicates that $\mathcal{L}_{rgb}$, $\mathcal{L}_{flow}$, and $\mathcal{L}_{depth}$ are all computed only within the foreground region defined by mask $\mathbf{M}_{i}$, ensuring that our optimization focuses on the target object.

% \textcolor{red}{We align the relative depth maps with the metric depth maps by computing a per-frame global scale and shift and use them for our optimization.} 

The tracking losses incorporate a 2D joint tracking term and a 2D vertex tracking term that leverage Cotracker3 \cite{karaev2024cotracker3} to trace selected keypoints throughout the video sequence. We project visible joints and vertices of the 3D static object onto image planes to establish keypoints for the first frame. These keypoints are then tracked through the entire video sequence using Cotracker3 to obtain $\mathbf{p}_{i}$, which represents the tracked positions at each frame $i$. Simultaneously, $\mathbf{p}_{i}^{\prime}$ is derived by projecting the deformed joints and mesh vertices onto image planes. The tracking losses are formulated as:

\begin{equation}
    \begin{aligned}
        &\mathcal{L}_{joint\_track}  = \sum_{i}{\left\|\mathbf{M}_{j} \odot (\mathbf{p}_{i, joint} - \mathbf{p}_{i, joint}^{\prime})\right\|^{2}},
    \\
&\mathcal{L}_{point\_track}  = \sum_{i}{\left\|\mathbf{M}_{v} \odot (\mathbf{p}_{i, vertex} - \mathbf{p}_{i, vertex}^{\prime})\right\|^{2}},
    \end{aligned}
\end{equation}
where $\mathbf{M}_{j}$ and $\mathbf{M}_{v}$ denote the visibility masks for joints and vertices in the first frame, respectively. These masks ensure that only visible keypoints contribute to the optimization process. Additionally, we also incorporate regularization terms that prevent temporal jittering by penalizing large transformation changes between consecutive frames. To balance these losses, we weight each term to ensure comparable magnitudes. In practice, regularization losses are down-weighted by 3–4 orders of magnitude relative to rendering and tracking losses to prevent over-smoothing.

Our animation optimization takes approximately 20 minutes for objects with up to 10K vertices on a single NVIDIA A100 GPU, processing 5-second videos (approximately 50 frames at 10 FPS) generated by Kling AI \cite{klingai} or JiMeng AI \cite{jimengai2025}. Runtime scales with both mesh complexity and frame count: (1) Mesh complexity: Models with more vertices will require additional Pytorch3D rendering time. For example, the bat case ($\sim$70K vertices) in our project page requires 90 minutes, while the turtle case ($\sim$15K vertices) takes 35 minutes. (2) Frame count: For a typical case taking 20 minutes at 50 frames (10 FPS, 5 seconds), increasing to 20 FPS (100 frames) extends optimization time to 41 minutes, while reducing to 4 FPS (20 frames) decreases it to 8 minutes, demonstrating approximately linear scaling with frame count.

After optimization, our animation process follows standard skeletal animation principles \cite{parent2012computer}: (1) Forward Kinematics (FK): We compute global joint transformations from optimized local transformations using hierarchical forward kinematics, traversing the skeleton from root to leaves; (2) Linear Blend Skinning (LBS) \cite{lewis2023pose}: Deform mesh vertices using weighted combinations of joint transformations, where each vertex is influenced by multiple joints according to skinning weights. This produces the final mesh animation sequence.

\subsection{Experimental details}

For baseline comparisons, we use the publicly available implementations of UniRig \cite{zhang2025unirig}, RigNet \cite{xu2020rignet}, and Pinocchio \cite{baran2007automatic} from their respective GitHub repositories. The Geodesic Voxel Binding (GVB) \cite{dionne2013geodesic} comparison utilizes the implementation in Autodesk Maya \cite{AutodeskMaya2024}. RigNet and MagicArticulate \cite{song2025magicarticulate} are trained on \ourdata{} using the original data preprocessing pipelines and training schedules specified by their authors.

\section{More details of \ourdata{}}
\label{detail_data}

\begin{figure*}
    \centering
\includegraphics[width=\textwidth]
{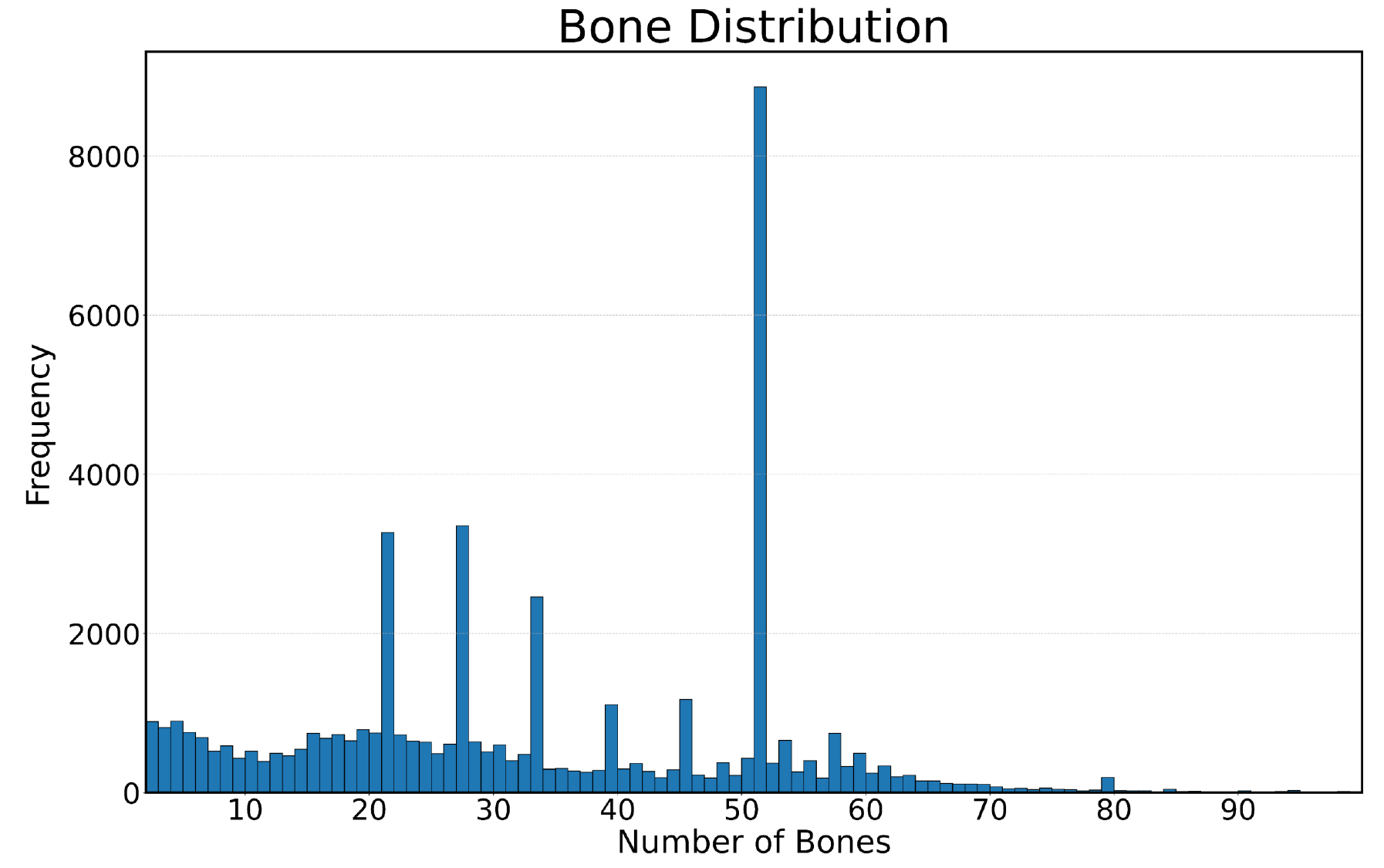}

    \caption{\textbf{Bone number distributions of \ourdata{}.} }
    \label{bone_num}
  \end{figure*}

Our dataset \ourdata{} is sourced from Objaverse-XL \cite{deitke2023objaverse, deitke2024objaverse}, specifically focusing on the GitHub and Sketchfab subsets that include file types containing rigging information (e.g., glb/gltf, fbx, dae, blend, etc.). From an initial 2.97M models, we extract 74K rigged assets (after removing over 150K duplicates) and curate 48K high-quality rigged models through quality verification. \Cref{xlv2} details these statistics. The number of bones in \ourdata{} ranges from 2 to 100, with the distribution illustrated in \Cref{bone_num}.

\begin{figure*}
    \centering
\includegraphics[width=\textwidth]
{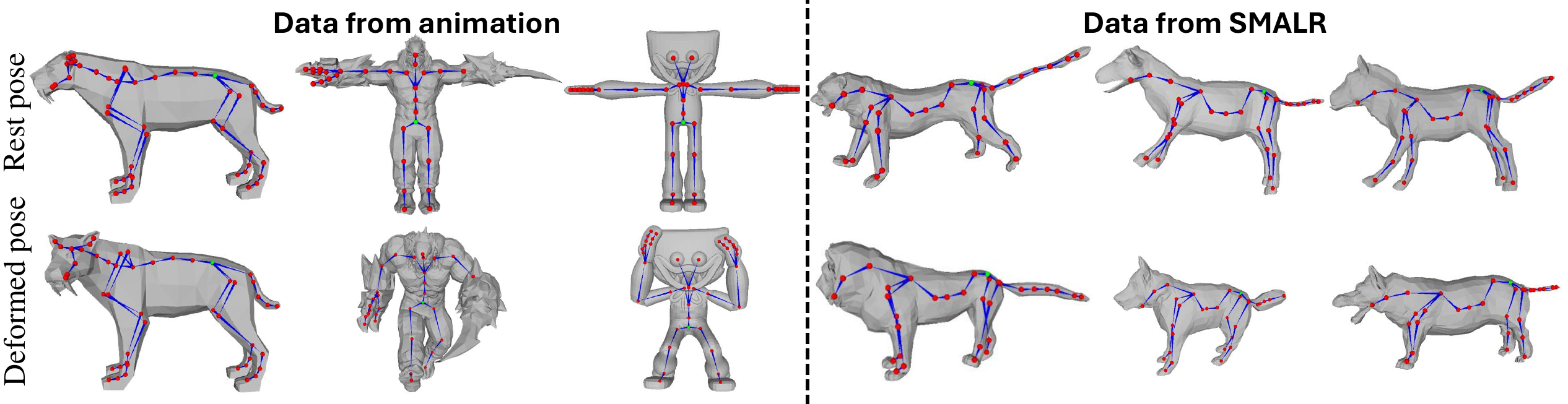}
    \caption{\textbf{Examples from the diverse-pose subset of \ourdata{}.} The left group showcases animation-derived samples: the top row displays the original rest poses, while the bottom row shows their corresponding deformed articulations extracted from animation sequences at frames of maximum pose deviation. The right group displays synthetically generated articulations created using SMALR \citep{Zuffi:CVPR:2017, Zuffi:CVPR:2018}.}
    \label{pose_data}
  \end{figure*}

\begin{figure*}
    \centering
\includegraphics[width=\textwidth]
{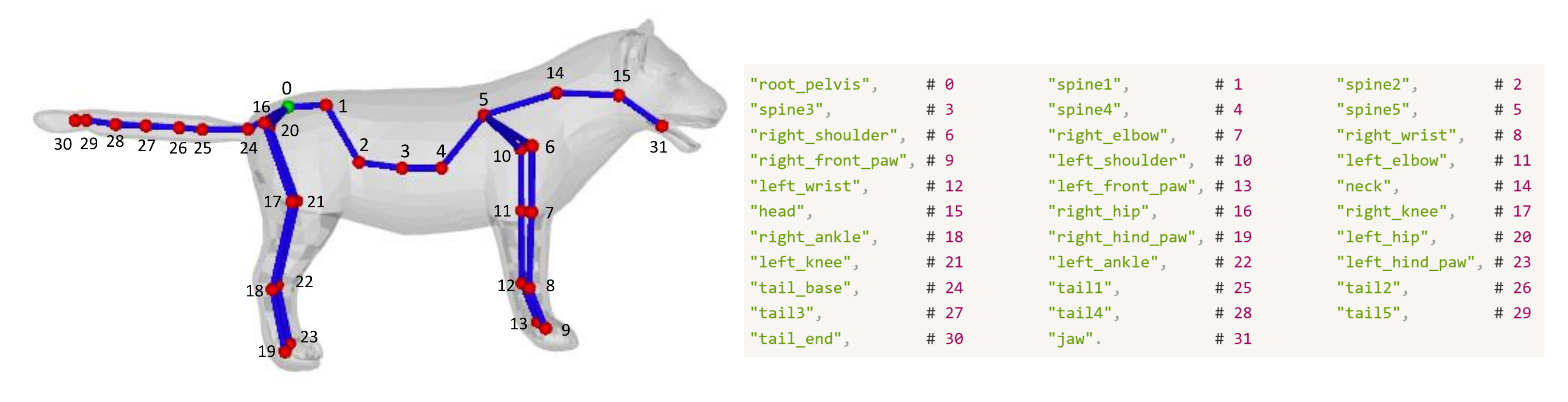}

    \caption{\textbf{Skeleton structure and joint names of SMALR data \cite{Zuffi:CVPR:2017, Zuffi:CVPR:2018}.} }
    \label{smal_skel}
  \end{figure*}

The diverse-pose subset in \ourdata{} is derived from two sources. The first component consists of poses extracted from data with animation, where we specifically selected frames exhibiting maximum deviation from rest pose configurations to capture extreme articulations. The second component comprises synthetically generated poses created using SMALR \citep{Zuffi:CVPR:2017, Zuffi:CVPR:2018} with parameterizations derived from 41 distinct animal scans and randomized valid poses. These randomized valid poses are generated by applying random rotation angles to SMALR animal joints while constraining angles within anatomically valid ranges, ensuring diverse articulation states while maintaining biological plausibility. Refer to \Cref{smal_skel} for the skeleton structure and joint names of SMALR data. There are originally two joints in index 0, we merge them into a single root joint.
Some examples from this diverse-pose collection are illustrated in \Cref{pose_data}. \textbf{Note that the models in the diverse-pose test set, along with their corresponding rest poses, are entirely excluded from the training data, ensuring a rigorous evaluation of generalization to both novel shapes and novel articulations.}

\begin{table*}[h]
% \vspace{-10pt}
  \caption{\textbf{Data statics for \ourdata{}.}}
  \label{xlv2}
  \centering
  \begin{tabular}{ccccc}
%   \hline
    \toprule
     Source &  All models  & with rigging & high-quality rigging & low-quality rigging \\
    \midrule
    Sketchfab  & 0.89M & 64K & 42K & 22K \\
    GitHub   & 2.08M & 10K  & 6K & 4K \\
     Total   & 2.97M & 74K &48K & 26K \\
    \bottomrule
  \end{tabular}
\end{table*}

\section{Additional experimental results}
\label{additioanl_results}
\subsection{More results of skeleton generation}

\begin{figure*}
    \centering
\includegraphics[width=\textwidth]
{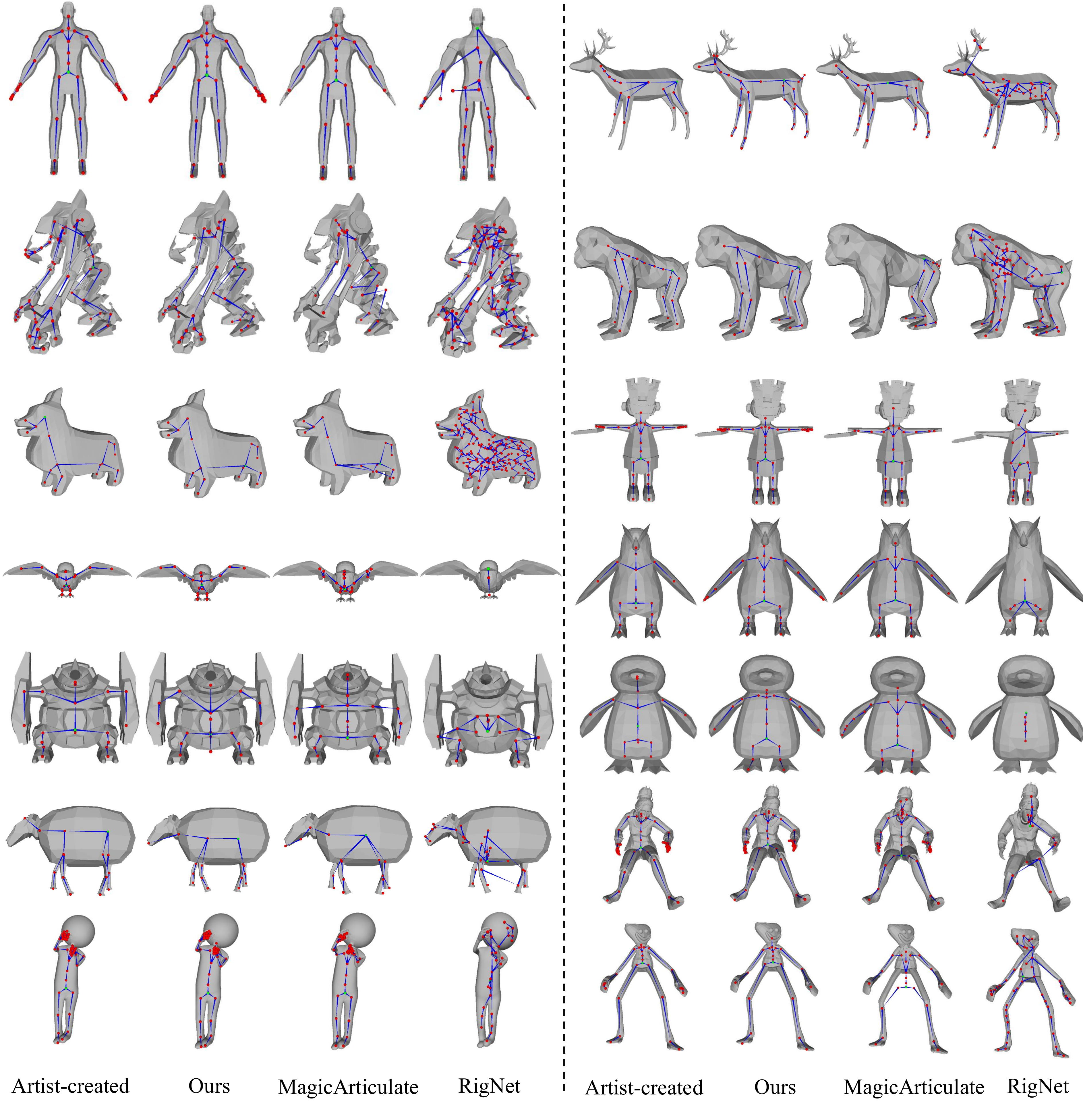}

    \caption{\textbf{Comparison of skeleton generation results on test sets.}
    From the top: six examples from \ourdata{}, four from \res{}, and four from the diverse-pose subset. Our method produces valid skeletons and even corrects artist-created errors (e.g., missing deer legs in row 1, missing penguin arms in rows 4–5).} 
    \label{supp_skel}
  \end{figure*}

\boldstartspace{More qualitative results on test sets.} \Cref{supp_skel} presents additional qualitative comparisons of skeleton generation by our method, MagicArticulate \cite{song2025magicarticulate}, and RigNet \cite{xu2020rignet} on meshes from \ourdata{}, \res{}, and the diverse-pose subset. Our method produces more valid skeletons and even corrects artist-created errors, such as the missing deer legs in the first row, and the missing penguin arms in rows 4 and 5.

\begin{figure*}
    \centering
\includegraphics[width=\textwidth]
{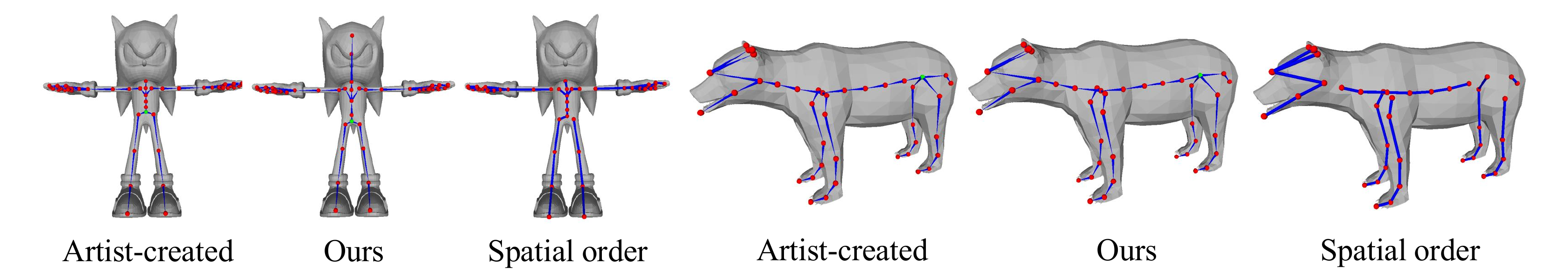}
\caption{\textbf{Comparison of sequence ordering.} Skeletons generated with hierarchical ordering (ours) remain connected, while spatial ordering always yields disconnected structures.} 
    \label{supp_order}
  \end{figure*}

\boldstartspace{Sequence ordering.} As a supplement to the ablation study in the main paper, \Cref{supp_order} compares skeletons generated with hierarchical and spatial ordering. Using spatial ordering always produces disconnected skeletons, as child joints generated before their parents create invalid parent references.

\boldstartspace{Inference time.} We compare average inference times on \ourdata{}-test (see \Cref{infer_skel}). Excluding data preprocessing for all methods, our approach is $2.6\times$ faster than Pinocchio \cite{baran2007automatic}, $3.0\times$ faster than RigNet \cite{xu2020rignet}, $1.9\times$ faster than UniRig \cite{zhang2025unirig}, and $1.6\times$ faster than MagicArticulate \cite{song2025magicarticulate}.

\begin{table*}[h]
% \vspace{-10pt}
  \caption{\textbf{Inference time of skeleton generation.}}
  \label{infer_skel}
  \centering
  \begin{tabular}{cccccc}
%   \hline
    \toprule
     Method &  Pinocchio  & RigNet & UniRig & MagicArticulate & Ours \\
    \midrule
     Inference time   & 3.9s & 4.5s &2.9s & 2.4s & 1.5s \\
    \bottomrule
  \end{tabular}
\end{table*}

\subsection{More results of skinning weight prediction}

\boldstartspace{Deformation error report.} In addition to the precision, recall, and L1-norm for evaluating skinning weight accuracy presented in the main paper, we also conducted a comprehensive assessment of practical efficacy through deformation error analysis. This supplementary metric quantifies the mean Euclidean distance between vertices deformed via predicted skinning weights and those deformed via ground truth weights across a diverse set of 10 randomly generated poses. As evidenced in \Cref{supp_skin_defor}, the proposed methodology exhibits better performance for all experimental datasets.

\begin{table*}[]
  \centering
  \caption{\textbf{Quantitative comparison of skinning weight prediction.} We evaluate our approach against GVB, RigNet, and MagicArticulate. For average L1-norm error (L1) and average distance error (avg Dist.), lower is better. Here, * denotes models trained on \ourdata{} with the diverse-pose subset.}
  \label{supp_skin_defor}
  \begin{tabular}{ccccccc}
    \toprule
    \multirow{2}{*}{Method} & \multicolumn{2}{c}{\ourdata{}} &  \multicolumn{2}{c}{\res{}} & \multicolumn{2}{c}{Diverse-pose} \\ \cmidrule{2-7}
  &  L1 $\downarrow$  &  avg Dist. $\downarrow$& L1 $\downarrow$  &  avg Dist. $\downarrow$ & L1 $\downarrow$  &  avg Dist. $\downarrow$  \\
    \midrule
     GVB  & 0.745  & 0.0087 & 0.687 & 0.0067
  & 0.786 & 0.0084 \\
    RigNet &  0.729 & 0.0082 & 0.707 & 0.0078 & 0.746 & 0.0089 \\
      MagicArti. & 0.451 & 0.0051 &  0.642
 & \underline{0.0064} &  0.479 &  0.0067 \\
      Ours   &  \underline{0.335} & \underline{0.0043} & \underline{0.443} & \textbf{0.0044}    & \underline{0.405} &  \underline{0.0061}  \\
      Ours*   &  \textbf{0.333} & \textbf{0.0042} & \textbf{0.442} & \textbf{0.0044}   & \textbf{0.353} &  \textbf{0.0053}   \\
    \bottomrule
  \end{tabular}
  % \vspace{-15pt}
\end{table*}

\boldstartspace{Ablation studies on block depth.} In addition to the ablation results in \Cref{ablation}, we also ablate the block depth. For the green attention block in \Cref{method}, we vary the number of stacked blocks, which we refer to as depth. When depth is 1 (44.3M parameters), this corresponds to a single block. We incrementally increase the block depth, corresponding to larger model size, and evaluate the resulting performance. The results are shown in \Cref{ab_depth}: when depth is set to 2 (87.7M parameters), performance on \ourdata{} and the diverse-pose subset improves noticeably, but there is a slight degradation on \res{}. When depth is 3 (130.9M parameters), performance on \ourdata{} and the diverse-pose subset shows comparable results to depth=2 but still exhibits a drop on \res{}. We attribute the drop on \res{} to the orientation distribution difference between \ourdata{} and \res{}: after training on \ourdata{} with highly varied orientations, the larger model may become biased toward diverse orientations, leading to degraded performance when testing on \res{}, which contains only front-facing orientations. For depth=3, the comparable performance to depth=2 without obvious improvement suggests that the model capacity may have reached saturation for the current dataset size.

\begin{table*}[]
  \centering
  \caption{\textbf{Ablation study on attention block depth in the skinning weight prediction network.}}
  \vspace{-5pt}
  \label{ab_depth}
  \begin{tabular}{cccccccccc}
    \toprule
    \multirow{2}{*}{Method} & \multicolumn{3}{c}{\ourdata{}} &  \multicolumn{3}{c}{\res{}} & \multicolumn{3}{c}{Diverse-pose} \\ \cmidrule{2-10}
 &  Prec. $\uparrow$ & Rec. $\uparrow$ &  L1 $\downarrow$&Prec. $\uparrow$ & Rec. $\uparrow$&  L1 $\downarrow$&Prec. $\uparrow$  & Rec. $\uparrow$&  L1 $\downarrow$  \\
    \midrule
depth = 1   &  87.6\% & \textbf{74.0\%} & 0.335
 & \underline{79.7\%} & \textbf{81.6\%} & \textbf{0.443}   & {83.6\%} & \textbf{72.2\%} & 0.405  \\
depth = 2& \underline{89.3\%} & \underline{73.0\%}
 & \textbf{0.316} & \textbf{79.8\%} & \underline{80.2\%} & \underline{0.453}  & \textbf{85.8\%} & \underline{71.1\%} & \underline{0.392} \\
depth = 3 & \textbf{89.4\%} &
72.5\% & \underline{0.317} & 79.6\% & 79.7\% & 
0.461 & \underline{85.7\%} & 70.8\% & \textbf{0.391} \\ 
    \bottomrule
  \end{tabular}
  \vspace{-10pt}
\end{table*}

\begin{figure*}
    \centering
\includegraphics[width=\textwidth]
{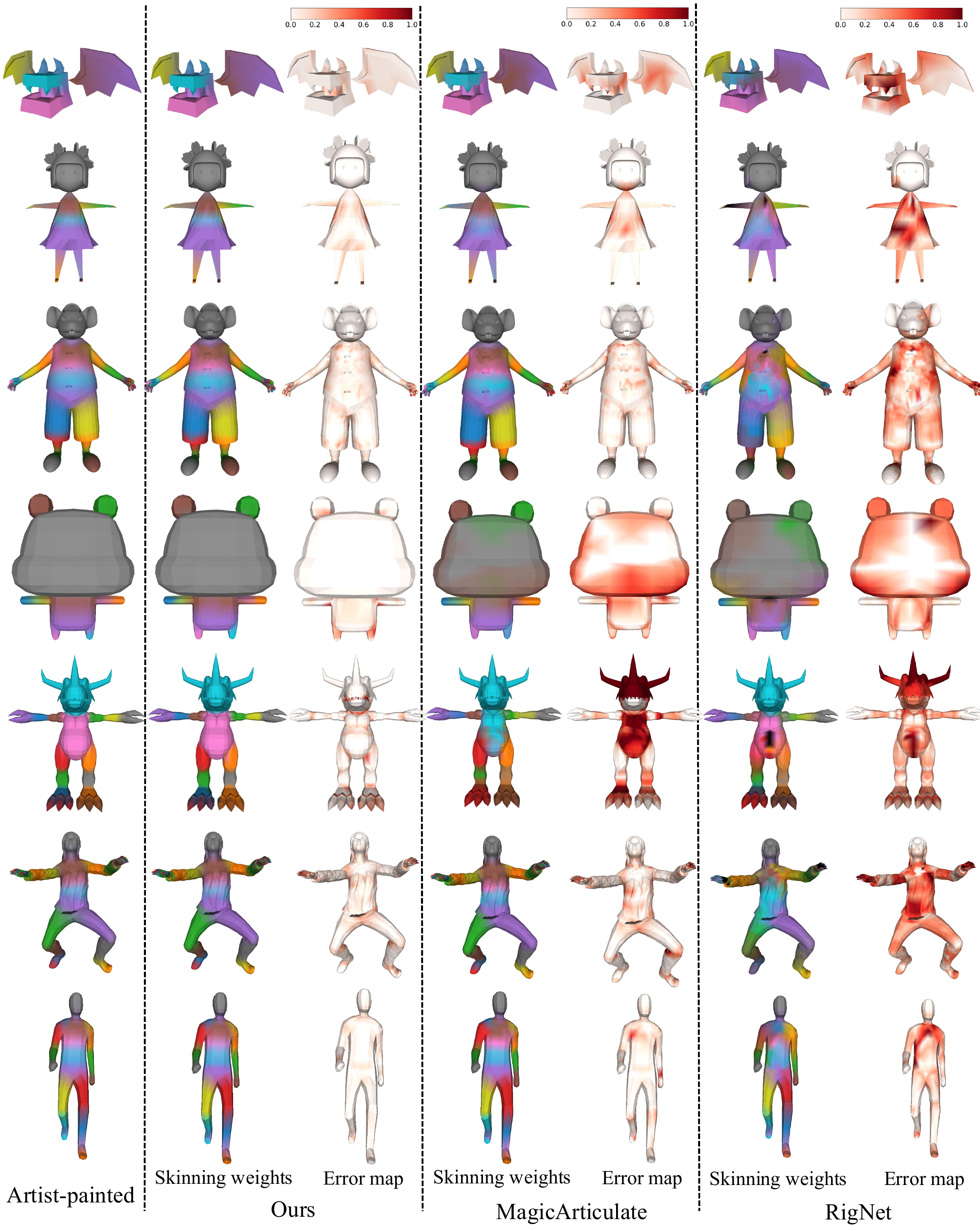}

    \caption{\textbf{Comparison of skinning weight prediction results.}
    From the top: three examples from \ourdata{}, two from \res{}, and two from the diverse-pose subset.
    Each pair shows the predicted weight visualization alongside its L1 error map. Our predictions more closely match the artist-painted references.} 
    \label{supp_skin}
  \end{figure*}

\boldstartspace{More qualitative results.} \Cref{supp_skin} presents additional qualitative comparisons of skinning weight predictions by our method, MagicArticulate \cite{song2025magicarticulate}, and RigNet \cite{xu2020rignet} on meshes from \ourdata{}, \res{}, and the diverse-pose subset. Each example pairs the predicted weight map with its L1 error map against artist-painted references, highlighting our method’s superior accuracy across diverse object categories.

\boldstartspace{Inference time.} We also compare average inference times for skinning weight prediction on \ourdata{}-test (see \Cref{infer_skin}). Excluding data preprocessing, our method is $59\times$ faster than GVB \cite{dionne2013geodesic}, $1.75\times$ faster than RigNet \cite{xu2020rignet}, and $45\times$ faster than MagicArticulate \cite{song2025magicarticulate}.

\begin{table*}[]
% \vspace{-10pt}
  \caption{\textbf{Inference time of skinning weight prediction.}}
  \label{infer_skin}
  \centering
  \begin{tabular}{ccccc}
%   \hline
    \toprule
     Method &  GVB  & RigNet & MagicArticulate & Ours \\
    \midrule
     Inference time   & 1.895s & 0.056s & 1.430s & 0.032s \\
    \bottomrule
  \end{tabular}
\end{table*}

\subsection{More results of animation}
\begin{figure*}
    \centering
\includegraphics[width=\textwidth]
{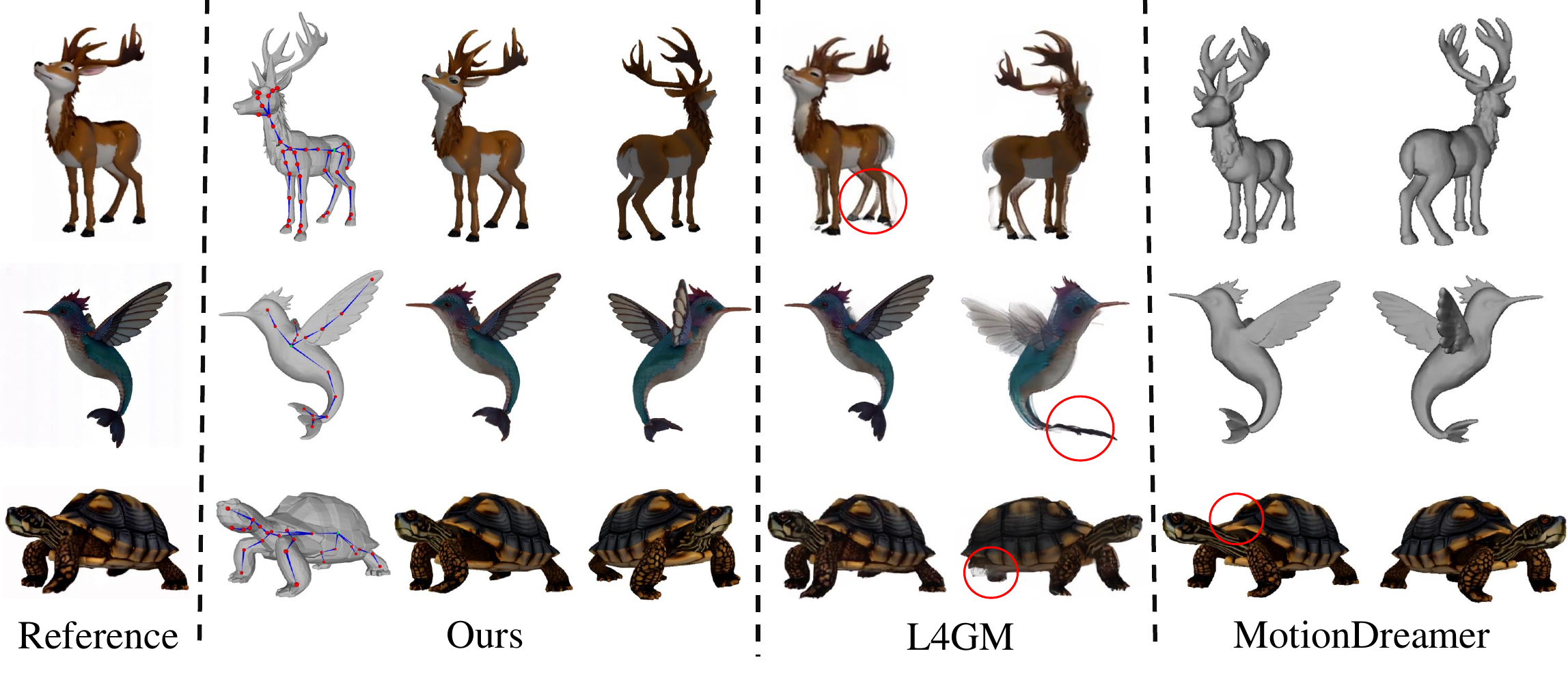}

    \caption{\textbf{Comparison of animation results on AI-generated meshes.} The meshes are generated by Tripo 2.0 \cite{tripo3d} and Hunyuan3D 2.0 \cite{hunyuan3d22025tencent}. \textbf{The shapes with skeletons represent the rest poses.}
    } 
    \label{supp_anime_ood}
  \end{figure*}

\boldstartspace{More qualitative results.} We present more animation results in \Cref{supp_anime_ood}. The input meshes are generated by Tripo 2.0 \cite{tripo3d}
 and Hunyuan3D 2.0 \cite{hunyuan3d22025tencent}. Despite well-aligned reference views, L4GM \cite{ren2024l4gm} consistently produces geometric distortions (highlighted in red), even with ground-truth multi-view renderings. MotionDreamer \cite{uzolas2025motiondreamer} generates subtle animations and introduces unintended deformations in rigid parts like the turtle shell. In contrast, our approach produces accurate, artifact-free animations with our generated rigging.
 
\boldstartspace{User study.} We conducted user studies with 21 participants to evaluate animation quality across three methods: L4GM \cite{ren2024l4gm}, MotionDreamer \cite{uzolas2025motiondreamer}, and our approach. Participants compared 8 animation examples across three evaluation criteria: (1) Video-Animation Alignment: Which animation result shows better alignment with the input video? (2) Motion Quality: Which one has a more natural and realistic motion? (3) 3D Geometry Preservation: Which method better maintains the original 3D object geometry without introducing distortions or artifacts? Results are shown in \Cref{user_study}. Our method outperforms both L4GM and MotionDreamer across all three evaluation dimensions. Note that Video-Animation Alignment is not evaluated for MotionDreamer since it uses text-driven motion generation rather than video guidance. 

\begin{table*}[h]
  \centering
  \vspace{-5pt}
  \label{user_study}
  \caption{\textbf{User study evaluation of animation results.}}
  \begin{tabular}{cccc}
    \toprule
    Method & Video-Animation Align. & Motion Quality & Geometry Preservation \\
    \midrule 
     MotionDreamer  & - &
0 & 0 \\
L4GM  & 19.64\%  & 16.67\% &
18.45\%  \\
      Ours   &  \textbf{80.36\%} &
 \textbf{83.33\%} &
\textbf{81.55\%}   \\
    \bottomrule
  \end{tabular}
  \vspace{-5pt}
\end{table*}

\section{Discussions}
\label{limit}
Despite its strong performance, our framework has two main limitations. First, it cannot capture fine-scale deformations, such as flowing hair or fluttering cloth, because no skeletons are generated for these highly deformable parts. A clear example is the swimming turtle sequence on our \href{https://chaoyuesong.github.io/Puppeteer/}{project page}. While the reference video shows very soft motion of the turtle's forelimbs, our animation results appear less smooth due to insufficient joint density in these regions. This limitation stems from our skeleton generation producing few joints in areas requiring fine-scale deformations. Animation-driven joint refinement could improve smoothness but remains future work. Second, the animation stage relies on per-scene optimization, which prevents real-time deployment. An end-to-end feed-forward model that predicts animation directly would eliminate this bottleneck.

Beyond these structural issues, several practical factors also affect animation quality. (1) Complex motion: Rapid movements with large joint rotations present challenges. For instance, in the seahorse case on our project page, while we capture the overall tail oscillation pattern, precise alignment with fine-scale movements remains difficult. Adaptive frame sampling based on optical flow magnitude—sampling more densely during rapid motion—could potentially address these challenges. (2) Video generation quality: Although current text-to-video models (Kling AI \cite{klingai}, JiMeng AI \cite{jimengai2025}) can generate complex motions with high success rates, video quality directly impacts animation fidelity. Motion blur or temporal inconsistencies degrade joint/vertex tracking accuracy and make optimization more challenging. We mitigate this by generating multiple video candidates and selecting the highest-quality one based on visual clarity and motion consistency. While this approach helps reduce the impact of poor-quality videos, video generation quality still represents a
limitation for highly complex motion scenarios. (3) Viewpoint and occlusion issues: Suboptimal camera angles can cause depth ambiguities and tracking failures. While we can select optimal viewpoints that maximize joint visibility using the input 3D mesh, single-view optimization inherently struggles when critical joints remain occluded throughout the sequence. Multi-view priors could provide better geometric understanding of occluded regions.

\section{Broader impact}
\label{impact}
Beyond its technical breakthroughs, this work holds significant societal implications by removing the need for specialized expertise and empowering a diverse range of creators who were once excluded from animation tools. As digital environments increasingly influence how we work, learn, and connect, expanding access to animated content creation becomes not only technically valuable but socially essential. However, democratizing animation technology also raises concerns about potential misuse in creating deceptive content and impacts on traditional animation industry employment. Our ultimate vision is to transform 3D animation from an exclusive professional skill into an intuitive creative medium accessible to everyone, while encouraging responsible use.

\bibliography{Reference}

\end{document}